\documentclass[]{smile}
\usepackage{xcolor}         
\definecolor{RowColor}{rgb}{0.93, 0.98, 0.97}
\definecolor{deepgray}{rgb}{0.5, 0.5, 0.5}
\definecolor{Salmon}{rgb}{1.0, 0.55, 0.41}
\definecolor{RoyalBlue}{rgb}{0.25, 0.41, 1.0}
\usepackage[utf8]{inputenc} 
\usepackage{url}            
\usepackage{hhline}
\usepackage{amsbsy,amsmath}
\usepackage{nicefrac}       
\usepackage{microtype}      
\usepackage{colortbl}
\usepackage{graphicx}
\usepackage{amsmath}
\usepackage{amssymb}
\usepackage{marvosym}
\usepackage{multirow}
\usepackage{makecell}
\usepackage{url}            
\usepackage{amsfonts}       
\usepackage{nicefrac}       
\usepackage{microtype}      
\usepackage{xcolor}         
\usepackage{enumitem}
\usepackage{graphicx}
\usepackage{multirow}
\usepackage{amsmath}
\usepackage{amstext}
\usepackage{MnSymbol}
\usepackage{wasysym}
\usepackage{listings}
\usepackage{color,colortbl}
\usepackage{wrapfig}
\usepackage{soul}
\usepackage{arydshln}
\usepackage{bm}
\usepackage{algorithm}
\usepackage{algpseudocode}
\usepackage{textcomp}
\usepackage{multirow}
\usepackage{subcaption}
\usepackage{tabularx}
\usepackage{adjustbox}
\usepackage{makecell}
\usepackage{array}
\usepackage{amssymb}
\usepackage{graphicx}

\usepackage{bxcoloremoji}

\usepackage{pifont}
\newcommand{\cmark}{\textcolor{green!60!black}{\ding{51}}}  
\newcommand{\xmark}{\textcolor{red!70!black}{\ding{55}}}    

\usepackage{CJKutf8}

\definecolor{qbg}{RGB}{210,230,250}
\definecolor{qrule}{RGB}{30,108,200}
\definecolor{gtbg}{RGB}{226,243,229}
\definecolor{gtrule}{RGB}{ 38,139, 60}
\definecolor{okbg}{RGB}{220,242,224}
\definecolor{okrule}{RGB}{ 38,139, 60}
\definecolor{badbg}{RGB}{253,228,228}
\definecolor{badrule}{RGB}{198, 60, 60}
\definecolor{warnbg}{RGB}{255,243,212}
\definecolor{warnrule}{RGB}{197,148, 18}

\newtcolorbox{questionbox}{enhanced, breakable, colback=white, colframe=qrule,
  boxrule=0.6pt, arc=2pt, left=8pt, right=8pt, top=4pt, bottom=4pt,
  title={\textbf{Task}},
  fonttitle=\bfseries\small, coltitle=white, colbacktitle=qrule}

\newtcolorbox{questionbox2}{enhanced, colback=white, colframe=qrule, boxrule=0.4pt, arc=2pt, left=6pt, right=6pt, top=3pt, bottom=3pt, title={\textbf{CoMT Example}}, fonttitle=\bfseries\small, coltitle=white, colbacktitle=qrule}

\newtcolorbox{questionbox3}{enhanced, colback=white, colframe=qrule, boxrule=0.4pt, arc=2pt, left=6pt, right=6pt, top=3pt, bottom=3pt, title={\textbf{BLINK-Jigsaw Example}}, fonttitle=\bfseries\small, coltitle=white, colbacktitle=qrule}

\newtcolorbox{questionbox4}{enhanced, colback=white, colframe=qrule, boxrule=0.4pt, arc=2pt, left=6pt, right=6pt, top=3pt, bottom=3pt, title={\textbf{Word-Search Example}}, fonttitle=\bfseries\small, coltitle=white, colbacktitle=qrule}

\newtcolorbox{okbox}[1]{enhanced, colback=white,    colframe=okrule,
  boxrule=0.4pt, arc=2pt, left=6pt, right=6pt, top=3pt, bottom=3pt,
  title={#1}, fonttitle=\bfseries\small, coltitle=white, colbacktitle=okrule,
  before upper={\parindent=0pt\leavevmode}}
\newtcolorbox{badbox}[1]{enhanced, colback=white,  colframe=badrule,
  boxrule=0.4pt, arc=2pt, left=6pt, right=6pt, top=3pt, bottom=3pt,
  title={#1}, fonttitle=\bfseries\small, coltitle=white, colbacktitle=badrule,
  before upper={\parindent=0pt\leavevmode}}
\newtcolorbox{warnbox}[1]{enhanced, colback=white, colframe=warnrule,
  boxrule=0.4pt, arc=2pt, left=6pt, right=6pt, top=3pt, bottom=3pt,
  title={#1}, fonttitle=\bfseries\small, coltitle=white, colbacktitle=warnrule,
  before upper={\parindent=0pt\leavevmode}}

\title{MUSE: A Unified Agentic Harness for MLLMs}

\author[*]{Jianglin Lu}
\author[*]{Hailing Wang}
\author[]{Xu Ma}
\author[]{Qihua Dong}
\author[]{Mingyuan Zhang}
\author[]{Yizhou Wang }
\author[]{Yun Fu}

\affiliation[]{Northeastern University}

\projectlinks{Code: \href{https://github.com/Jianglin954/MUSE}{https://github.com/Jianglin954/MUSE}
}

\contribution[*]{Equal Contribution. Corresponding author: \email{jianglinlu@outlook.com}}

\abstract{
\textbf{\large Abstract} Despite rapid progress, multimodal large language models (MLLMs) still fail on tasks that humans solve effortlessly, such as navigating a grid maze from a screenshot or selecting the correct puzzle piece.
Rather than retraining the model, we ask a complementary question: {how much capability can be elicited from a frozen MLLM purely by improving the execution scaffold around it?}
We introduce MUSE, a multimodal unified structured execution harness that wraps any off-the-shelf MLLM with composable modules for task representation, visual processing, perception tool use, structured parsing, deterministic verification, and verifier-guided repair, without any model retraining.
We evaluate MUSE across diverse benchmarks spanning visual spatial planning, visual perception, multimodal reasoning, and fine-grained visual discrimination, using multiple state-of-the-art MLLMs.
MUSE delivers consistent gains over the bare model in all settings, with the largest jumps on challenging instances.
Further analysis reveals that many MLLM failures arise from harness-level shortcomings rather than fundamental model deficits, and can be addressed through verifier-guided repair without touching the model.
These findings highlight the agentic multimodal harness as a critical yet underexplored design dimension, offering an orthogonal avenue for improving MLLMs beyond model-centric optimization.
}

\begin{document}

\maketitle

\section{Introduction}
\label{sec:intro}

Multimodal large language models (MLLMs)~\citep{liu2023llava, openai2024gpt4o, openai2024gpt4, claude3} extend large language models (LLMs)~\cite{llama3herdmodels, qwen, GPT3} with visual perception, unifying image understanding, dialogue, and reasoning within a single interface and achieving impressive performance on various vision tasks.
Despite this rapid progress, MLLMs still exhibit striking failures on tasks that humans solve almost instantaneously.
Examples include identifying which puzzle piece completes a missing corner~\citep{blink2024}, counting non-overlapping objects~\citep{tong2024mmvp}, or planning a valid path through a simple grid maze~\citep{vsp2025}.
These failures persist despite the strong general capabilities of modern MLLMs and pose a major challenge for applications requiring reliable visual cues, such as vision reasoning~\citep{li2026latent, dong2026refadv, yang2025machine}, multimodal understanding~\citep{yue2025mmmupro, lu2026indra}, and embodied control~\citep{liu2025visualagentbench, zhang2026thinkjepa, OOSEmbodiedAgents}.

A dominant direction in recent research has been to improve the model itself.
Existing approaches introduce synthetic visual reasoning data~\citep{sat2025, wang2025monet, yang2025machine}, visual chain-of-thought supervision~\citep{cotvla2025,zhang2023mmcot, dong2026cotreferring}, or reinforcement-learning-based self-verification mechanisms~\citep{neurips2025aha,verifierselfvla2026, lu2026restorer1}.
While effective, these approaches fundamentally rely on model-centric optimization, requiring either access to model parameters or expensive retraining pipelines.
In practice, however, the strongest commercial MLLMs are closed-source, and even open-weight models evolve rapidly, making specialized training pipelines costly to maintain and quickly outdated as new checkpoints emerge.
Therefore, we argue for a complementary and orthogonal direction: \textit{improving the harness surrounding the model rather than the model itself}~\citep{Meng2026agentharness, tang2026agent, ning2026codeagentharness, li2026agentharness}.

Recent studies on LLM agents suggest that many practical failures arise not from insufficient pretraining, but from weaknesses in the surrounding execution framework, or ``exoskeleton'', which governs information organization, tool invocation, output verification, and retry strategies~\citep{wang2026heavyskill, lin2026museautoskill, lin2026agenticharness}.
Consequently, a growing body of work has shifted attention from model optimization to harness engineering.
For example, Meta-Harness~\citep{lee2026metaharness} treats harness engineering as an optimization problem, where an LLM iteratively improves its own execution logic by analyzing past execution traces and rewriting the surrounding harness code.
AutoHarness~\citep{lou2026autoharness} uses LLMs to synthesize and refine executable harnesses using environment feedback with tree-based code search.
Life-Harness~\citep{xu2026adaptinginterface} introduces a model-agnostic runtime interface that adapts the harness rather than the model, converting recurring agent-environment failures into reusable interventions.
\textit{However, harness strategies for MLLMs remain largely unexplored, despite identical economic incentives.}

Transferring the harness paradigm from text-only tasks to multimodal settings is non-trivial: visual inputs introduce perceptual brittleness, outputs might require verification against geometric or spatial structure, and tool invocation must operate over visual context rather than disembodied text.
To address these challenges, we introduce {MUSE}, a \textbf{multimodal unified structured execution harness} for MLLMs.
Rather than modifying the model itself, MUSE treats the surrounding execution framework as a first-class optimization target.
It wraps a frozen MLLM with a modular inference pipeline that covers the full inference path from perception and tool use, through structured parsing and deterministic verification, to verifier-guided repair, all under a unified task-level abstraction.
Each component is explicitly designed to address specific failure modes observed in multimodal tasks.
Importantly, MUSE operates under a strict black-box setting: the underlying model is never fine-tuned or altered, and all performance gains arise purely from harness-level improvements.

We evaluate the proposed MUSE on four heterogeneous benchmarks that stress different aspects of multimodal applications: visual spatial planning (VSP-Grid~\citep{vsp2025}), visual perception (BLINK-Jigsaw ~\citep{blink2024}), multimodal reasoning (CoMT~\citep{comt2025}), and fine-grained visual discrimination (the Word Search sub-task of TIR-Bench~\citep{li2025tir}).
Across four frontier models (GPT-4o~\citep{openai2024gpt4o}, GPT-5.4~\citep{singh2026openaigpt5card}, Claude Haiku~4.5~\citep{claudehaiku}, and Claude Opus~4.7~\citep{claudeopus}), MUSE consistently improves performance without any model tuning.
The improvements are particularly pronounced on challenging instances, where repair fixes a large fraction of otherwise incorrect predictions. 
Detailed analyses attribute these gains to corrections of diverse failure modes, including perceptual errors, planning inconsistencies, and output-format violations, while also exposing remaining bottlenecks such as persistent mis-localization in dense visual scenes.
Our main contributions are summarized as follows:
\begin{itemize}[leftmargin=15pt]

    \item We propose {MUSE}, a unified agentic harness that improves frozen MLLMs by optimizing the execution scaffold rather than the model itself. To the best of our knowledge, \textit{MUSE is the first unified multimodal harness designed specifically for frozen MLLMs on visual-centric tasks}.

    \item We introduce a modular inference pipeline that unifies task representation, visual preprocessing, structured parsing, deterministic verification, and iterative repair under a single abstraction.

    \item We evaluate MUSE on multiple heterogeneous multimodal benchmarks across several state-of-the-art MLLMs, and demonstrate consistent improvements, with the largest gains on challenging instances, e.g., GPT-4o on Word Search improves from $3\%$ to $21\%$. Our failure-mode analysis shows that many MLLM failures arise from harness-level shortcomings rather than fundamental model deficits. 

\end{itemize}

\section{Related Work}
\label{sec:related}

\paragraph{MLLMs.}
MLLMs extend the autoregressive language-modeling paradigm to multimodal inputs by aligning a pretrained vision encoder with an LLM backbone, followed by large-scale multimodal instruction tuning~\citep{liu2023llava, openai2024gpt4o, claude3, lu2025representation}.
They ingest interleaved text and image inputs and produce fluent natural-language responses, achieving competitive performance on various vision tasks~\citep{li2026latent, dong2026visualreason,  liu2025visualagentbench, zhang2025vqtoken}.
Yet despite this progress, MLLMs remain unreliable on tasks that humans solve effortlessly, such as object counting~\citep{tong2024mmvp}, spatial planning~\citep{vsp2025}, puzzle piece completion~\citep{blink2024}, or reading dense character grids~\citep{li2025tir}.
One sub-line of MLLM research attempts to compensate for these gaps by augmenting the model with external tools. 
For example, MM-ReAct~\citep{yang2023mmreact}, Visual ChatGPT~\citep{wu2023visualchatgpt}, HuggingGPT~\citep{shen2023hugginggpt}, ViperGPT~\citep{suris2023vipergpt}, and VisProg~\citep{gupta2023visprog} decompose complex queries into calls to specialized vision modules, and Set-of-Mark prompting~\citep{yang2023setofmark} uses SAM~\citep{kirillov2023sam} to annotate image regions with symbolic tags fed back to the MLLM.
We treat the frozen MLLM as the perception engine and focus optimization on the harness layer, where MUSE can selectively invoke perception tools through a registry to enhance visual reasoning and task execution.

\paragraph{LLM Agents.}
LLM-based agents extend single-shot generation into multi-step decision processes that combine reasoning, tool invocation, and feedback in a structured loop, with ReAct~\citep{yao2023react} popularizing the reasoning-action interleaving template and Toolformer~\citep{schick2023toolformer} showing that an LLM can learn when to invoke external APIs.
A prominent sub-line explores iterative self-improvement, where the model critiques its own outputs.
For example, Self-Refine~\citep{madaan2023selfrefine} uses the model to generate feedback and revise its own answer, Reflexion~\citep{shinn2023reflexion} accumulates episodic lessons across attempts, and Tree-of-Thoughts~\citep{yao2024tot} explores multiple reasoning branches with self-evaluation-based pruning.
A common limitation of self-critique is the self-verification gap: imperfect internal evaluation can miss errors that the same model also makes during generation~\citep{neurips2025aha}.
MUSE is positioned within the agent family but departs from self-critique by employing an externalized verifier whenever the task admits one.
This helps decouple generation from evaluation, providing more reliable refinement signals and mitigating the limitations of self-critique.

\paragraph{Harness Optimization.}
Treating the software scaffolding around a frozen LLM as the optimization variable has emerged as a productive alternative to model training itself.
Early work focuses on prompt optimization~\citep{yang2024opro,agrawal2025gepa}, and on text-based credit assignment for entire LLM systems~\citep{yuksekgonul2024textgrad, lee2025feedbackdescent}.
Declarative LLM pipelines~\citep{khattab2024dspy} further expose LLM workflows as structured programs, allowing prompts, demonstrations, and model parameters across multiple calls to be optimized jointly.
A recent line of work treats the \textit{harness}, the runtime scaffold that governs input construction, tool invocation, output verification, and retry strategies, as an explicit object of optimization rather than a static execution layer~\citep{openai2026harness, anthropic2025harness, boluk2026harness, young2025anthropic, bockeler2026harness}. 
For example, Meta-Harness~\citep{lee2026metaharness} formulates harness design as an outer-loop coding-agent optimization problem. 
AutoHarness~\citep{lou2026autoharness} automatically synthesizes and refines executable harnesses using environment feedback, and Life-Harness~\citep{xu2026adaptinginterface} introduces reusable runtime interventions that adapt agents to changing environments without finetuning.
Despite this progress, all the methods above are evaluated exclusively on text-only tasks, optimizing prompts, pipelines, or harnesses over purely textual artifacts. In contrast, we consider a multimodal setting in which the harness must reason over both textual and \textit{visual} signals.
Instead of searching over harnesses for text-only tasks, our proposed MUSE is designed as a runtime harness for multimodal, visually grounded benchmarks and applications. This shifts the focus from harness search to harness adaptation under multimodal constraints.

\section{The Unified Agentic Harness}
\label{sec:framework}

\subsection{Problem Statement}
\label{sec:framework:problem}
Following the harness-optimization framework of~\citep{lee2026metaharness}, we model the harness as a stateful program $H$ that wraps a frozen MLLM $\pi_\theta$.
Given a task instance $x \sim \mathcal{X}$, executing $H$ with $\pi_\theta$ induces a rollout trajectory $\tau \sim p_{\pi_\theta}(H, x)$, consisting of an initial model response followed by at most $K$ verifier-guided repair steps.
Each trajectory is evaluated by a task-specific deterministic reward $r(\tau, x)$, indicating success (e.g., whether a generated plan reaches the goal in a simulator, or whether a predicted answer matches the ground truth).
With $\pi_\theta$ fixed, the objective is to optimize the harness:
\begin{equation}
H^{\star} \;=\; \arg\max_{H \in \mathcal{H}} \;
\mathbb{E}_{x \sim \mathcal{X},\, \tau \sim p_{\pi_\theta}(H, x)}
\bigl[ r(\tau, x) \bigr],
\quad \text{subject to} \quad \mathbb{E}[c(\tau)] \le \kappa,
\label{eq:objective}
\end{equation}
where $c(\tau)$ denotes the execution cost and $\kappa$ is a per-task budget.

Two distinctions are critical for our setting.
First, each $x$ may include both text and images.
As a result, the harness space $\mathcal{H}$ must encompass not only textual operations but also image-dependent components, such as preprocessing, perceptual tool invocation, and image-grounded verification.
Second, unlike Meta-Harness, which searches over $\mathcal{H}$ using an outer-loop optimization procedure, we pursue a complementary design-driven approach.
Specifically, we construct a single harness $H \in \mathcal{H}$ by systematically addressing empirically observed failure modes in prior multimodal benchmarks~\citep{vsp2025,blink2024,comt2025,li2025tir}.

\subsection{Overview}
\label{sec:framework:overview}
MUSE is a multi-turn controller built around a frozen MLLM $\pi_\theta$, which is accessed solely through a black-box interface $\pi_\theta(\mathcal{M}) \mapsto \textit{text}$ over message lists $\mathcal{M}$.
Given a benchmark instance $x$ (consisting of a question, associated images, and optional metadata), MUSE executes a finite interaction transcript $(m_0,o_0,p_0,v_0),\ldots,(m_k,o_k,p_k,v_k)$ with $k \le K$, where $m_i$ denotes the constructed message context, $o_i$ the raw model output, $p_i$ a task-specific candidate parsed from $o_i$, and $v_i$ the verifier's verdict on $(o_i, p_i)$.
At termination, MUSE returns a structured result record containing the final success indicator, parsed outputs, intermediate textual traces, raw model responses, optional reference annotations, and execution metadata.
Importantly, MUSE operates without modifying $\pi_\theta$: no fine-tuning, distillation, or parameter updates are performed. All adaptation is realized through the harness $H$.
Algorithm~\ref{alg:harness} sketches the overall workflow of MUSE.
The remainder of this section introduces the key components of MUSE.

\begin{algorithm}[!tb]
\caption{MUSE: a multimodal unified structured execution harness for MLLMs.}
\label{alg:harness}
\begin{algorithmic}[1]
\Require Frozen MLLM $\pi_\theta$, task instance $x$, max repairs $K \ge 0$
\Ensure An \textsc{ApproachResult} indicating success/failure and the trajectory
\State $r \gets \textsc{TaskHooks}(x).\textsc{tools}$                     \Comment{Tool requirements from the task-type hook registry}
\State $\mathcal{T} \gets \textsc{RunRequiredTools}(r, x)$               \Comment{Run each declared perception tool on $x$'s image}
\State $m_0 \gets \textsc{BuildInitialPrompt}(x, \mathcal{T})$           \Comment{Optionally folds tool outputs into the prompt}
\State $o_0 \gets \pi_\theta(\textsc{Pack}(x, m_0))$                     \Comment{Single call to the frozen MLLM}
\State $p_0 \gets \textsc{Parse}(x, o_0)$                                \Comment{Task-specific parser extracts the structured candidate}
\State $v_0 \gets \textsc{Verify}(x, o_0, p_0, r, \mathcal{T})$          \Comment{Outcome policy plus perception-evidence gate}
\If{$v_0.\text{ok}$} \Return \textsc{ApproachResult(success=True, \dots)}  \Comment{Early-stop on first verifier acceptance}
\EndIf
\State $\mathcal{F} \gets [(o_0, v_0.\text{detail})]$                    \Comment{Initialize the failure transcript}
\For{$i = 1$ \textbf{to} $K$}                                            \Comment{Up to $K$ verifier-guided repair turns}
\State $m_i \gets \textsc{BuildRepairPrompt}(x, m_0, \mathcal{F}, i, K)$ \Comment{Adds the compressed failure summary}
\State $o_i \gets \pi_\theta(\textsc{Pack}(x, m_i))$                     \Comment{Re-query the frozen MLLM on the extended context}
\State $p_i \gets \textsc{Parse}(x, o_i)$                                \Comment{Re-parse the new response}
\State $v_i \gets \textsc{Verify}(x, o_i, p_i, r, \mathcal{T})$          \Comment{Re-verify under the same outcome policy}
\If{$v_i.\text{ok}$} \Return \textsc{ApproachResult(success=True, \dots)}  \Comment{Loop short-circuits for acceptance}
\EndIf
\State Append $(o_i, v_i.\text{detail})$ to $\mathcal{F}$                \Comment{Otherwise extend the failure transcript}
\EndFor
\State \Return \textsc{ApproachResult(success=False, \dots)}               \Comment{Budget exhausted without acceptance}
\end{algorithmic}
\end{algorithm}

\subsection{Core Components}
\label{sec:framework:components}

\paragraph{Unified Task Representation.}
To avoid dataset-specific branching in the execution loop, we represent each benchmark instance as a unified record $x$ produced by a per-dataset adapter.
Each adapter normalizes its source's on-disk format into the same record schema, so the executor sees a uniform iterator regardless of source.
Concretely, the executor consumes a single immutable record that encapsulates the task type, shared instructions, ordered image paths, optional structured inputs, and, when available, ground-truth annotations.
Task-specific behavior is handled separately through the abstractions described next.

\paragraph{Unified Task Hooks.}
Task hooks provide the concrete behavioral interface for each task type, binding the unified record schema to task-specific parsing, prompting, and repair logic.
For each task type, we define an immutable task hook record with four components:
i) {answer extraction}, which maps raw model outputs to the structured representation consumed by the outcome policy (e.g., handling boxed answers, partial JSON recovery, or direct QA pass-through);
ii) {instruction initialization}, which optionally overrides the default first-turn prompt with task-specific instructions;
iii) {repair policies}, which determine how verifier failures are translated into feedback for subsequent refinement; and
iv) {tool requirements}, which declare the perception tools the harness might invoke for this task.
This abstraction decouples task-specific logic from the execution loop, so that supporting a new benchmark reduces to registering an adapter and a task hook, without modifying the core executor.

\paragraph{Instruction Interface.}
The instruction interface constructs both the initial user message and subsequent repair prompts by composing dataset text, optional tool outputs, and references to preprocessed visual artifacts under the task-specific template registered in the task hook.
For repair turns, the interface appends an ordered transcript of prior model outputs and their verifier feedback to the original instruction, formatting each entry according to the repair-policy hook for the current task type.
By separating message assembly from template content and failure phrasing, the interface allows task-specific prompt wording to be updated without changing the assembly logic.

\paragraph{Outcome Policy.}
The outcome policy is the authority for determining solution correctness, externalizing task semantics from the model into the harness.
Verification is implemented as a family of task-specific verifiers, each encoding a notion of correctness aligned with the response schema, and optionally augmented with tool-auxiliary checks when annotations alone are insufficient to resolve perceptual ambiguity.
Each rejection is categorized along a coarse taxonomy capturing distinct failure modes, including format violations, task-logic errors, and evidence gaps.
The repair policy uses this taxonomy to select task-appropriate feedback phrasing for subsequent repair turns.
See Appendix~\ref{appendx_outcom_police} for the implementation details.

\paragraph{Simulator-Backed Verification.}
For tasks that admit executable environments, we instantiate verification via explicit simulators that provide grounded feedback.
For example, on navigation benchmarks, the verifier evaluates a candidate action sequence by simulating its execution.
Given the instance-specific transition dynamics, each action sequence induces a unique trajectory, and success is defined as reaching a designated goal state.
Failure cases yield localized signals (e.g., collisions, invalid transitions, or premature termination).
On jigsaw-style piece-fitting tasks, the verifier stitches each candidate piece into the target image and scores the resulting seams by the aggregate $L_1$ pixel mismatch, allowing it to differentiate candidates that are visually similar but exhibit different boundary consistency.

\paragraph{Perception Tool Registry.}
Beyond the core pipeline above, MUSE exposes an extensible registry of pluggable perception tools that can be invoked deterministically by the harness whenever a task benefits from auxiliary visual processing.
Each tool is a pure function from an image to a structured payload (text, embeddings, or another image), and is referenced by name from each task type's per-task tool requirements.
The executor runs only the tools that the current task declares.
Built-in tools include geometric transforms (e.g., zoom, rotation), OCR, and embedding utilities.
For example, on the Word Search task (Section~\ref{sec:word_search}), we register OCR as the perception tool and demonstrate that it composes cleanly with verifier-guided repair to address a perceptual bottleneck the single-shot baseline cannot resolve.

\paragraph{Repair Loop.}
When the verifier rejects the initial candidate, the harness enters a repair loop: each round re-invokes $\pi_\theta$ with the original message context augmented by the running failure transcript, re-parses the structured output, and re-verifies it until either acceptance or the repair budget of $K$ additional turns is exhausted.
The verifier decides when the loop terminates: the moment it returns success, the loop exits immediately, so most easy instances incur only the cost of the initial call.
The transcript is formatted to make failure modes explicit: attempts are sequentially indexed, model outputs are quoted, and simulator-derived feedback is clearly distinguished from syntactic or formatting errors.
On top of the raw transcript, the harness emits a compact failure summary that includes the count of each error type, the most recent verifier signal, and a repeated-output flag derived from a content fingerprint of the previous attempt.
This prevents the model from cycling on the same wrong commitment when feedback is intentionally short.

\section{Experiments}
\subsection{Experimental Setup}
\label{sec:setup}

\paragraph{Benchmarks.}
We evaluate MUSE on four heterogeneous vision benchmarks: VSP-Grid~\citep{vsp2025}, BLINK-Jigsaw~\citep{blink2024}, CoMT~\citep{comt2025}, and TIR-Bench~\citep{li2025tir}, which together cover visual spatial planning, visual perception, multimodal reasoning, and fine-grained visual discrimination.
{VSP-Grid} presents a rendered grid environment containing a player, a goal, and hazardous regions.
The task requires the model to generate a sequence of moves that reaches the goal without entering hazards.
Predicted sequences are parsed and executed in a simulator, and success is determined by whether the goal is reached; optimality is additionally assessed via path length.
We report results across four difficulty levels, corresponding to $n \times n$ grids with $n \in \{3,4,5,6\}$.
{BLINK-Jigsaw} provides three images: an incomplete puzzle with a missing region and two candidate patches.
The task is to choose the correct patch that correctly completes the image and output the answer in a prescribed format.
For CoMT, we use the creation split, which consists of multimodal geometry reasoning problems, each including a natural-language question, multiple-choice options, and one associated image.
The model must identify the single correct option.
We further evaluate the Word Search split of TIR-Bench, which provides a dense grid image populated with many similar glyphs (Chinese characters, Latin letters, or digits) and asks the model either to count the occurrences of a designated target glyph, or to localize a target glyph by its row and column indices.

\begin{table}[!tb]
\caption{Performance on the VSP-Grid task. V: Valid (\%), plans reaching the goal; O: Optimal (\%), plans with shortest-path length; Avg.: average across task levels.}
\label{table1_vsp}
\small
\centering
\setlength{\tabcolsep}{3.3mm}{
\begin{tabular}{l|c|cc|cc|cc|cc|cc}
\toprule
\multirow{2}{*}{Model} & \multirow{2}{*}{Approach} & \multicolumn{2}{c|}{Level 3} & \multicolumn{2}{c|}{Level 4} & \multicolumn{2}{c|}{Level 5} & \multicolumn{2}{c|}{Level 6} & \multicolumn{2}{c}{Avg.} \\
\cmidrule{3-12}
& & V & O & V & O & V & O & V & O & V & O \\
\midrule
\multirow{2}{*}{{GPT-4o}} 
& \texttt{BASE} & 0.65  & 0.41 & 0.46 & 0.34 & 0.39 & 0.26 & 0.23  & 0.16 & 0.43 & 0.29 \\
& \texttt{MUSE}
& 0.89 & {0.55} & {0.61} & {0.42} & {0.55} & {0.39} & {0.40} & {0.26} & {0.61} & {0.41} \\
\midrule
\multirow{2}{*}{{GPT-5.4}}
& \texttt{BASE} & 0.50  & 0.42 & 0.38 & 0.20 & 0.34 & 0.24 & 0.23  & 0.11 & 0.36 & 0.24 \\
& \texttt{MUSE} & 
0.66 & {0.53} & {0.67} & {0.39} & {0.65} & {0.43} & {0.37} & {0.18} & {0.59} & {0.38}\\
\midrule
\multirow{2}{*}{{Haiku 4.5}}
& \texttt{BASE} & 0.47 & 0.43  & 0.37 & 0.37 & 0.50 & 0.49 & 0.31  & 0.28 & 0.41 & 0.39 \\
& \texttt{MUSE} & 0.75 & {0.70} & {0.65} & {0.61} & {0.67} & {0.64} & {0.54} & {0.47} & {0.65} & {0.61}\\
\midrule
\multirow{2}{*}{{Opus 4.7}}
& \texttt{BASE} & 0.82 & 0.66  & 0.74 & 0.69 & 0.85 & 0.79 & 0.59  & 0.58 & 0.75 & 0.68 \\
& \texttt{MUSE} & 0.97 & {0.77} & {0.94} & {0.90} & {0.95} & {0.86} & {0.81} & {0.78} & {0.92} & {0.83}\\
\bottomrule
\end{tabular}}
\end{table}
\begin{figure*}[!tb]
  \centering
  \begin{subfigure}{0.245\linewidth}
  \includegraphics[scale=0.275,trim=155 0 155 0,clip]{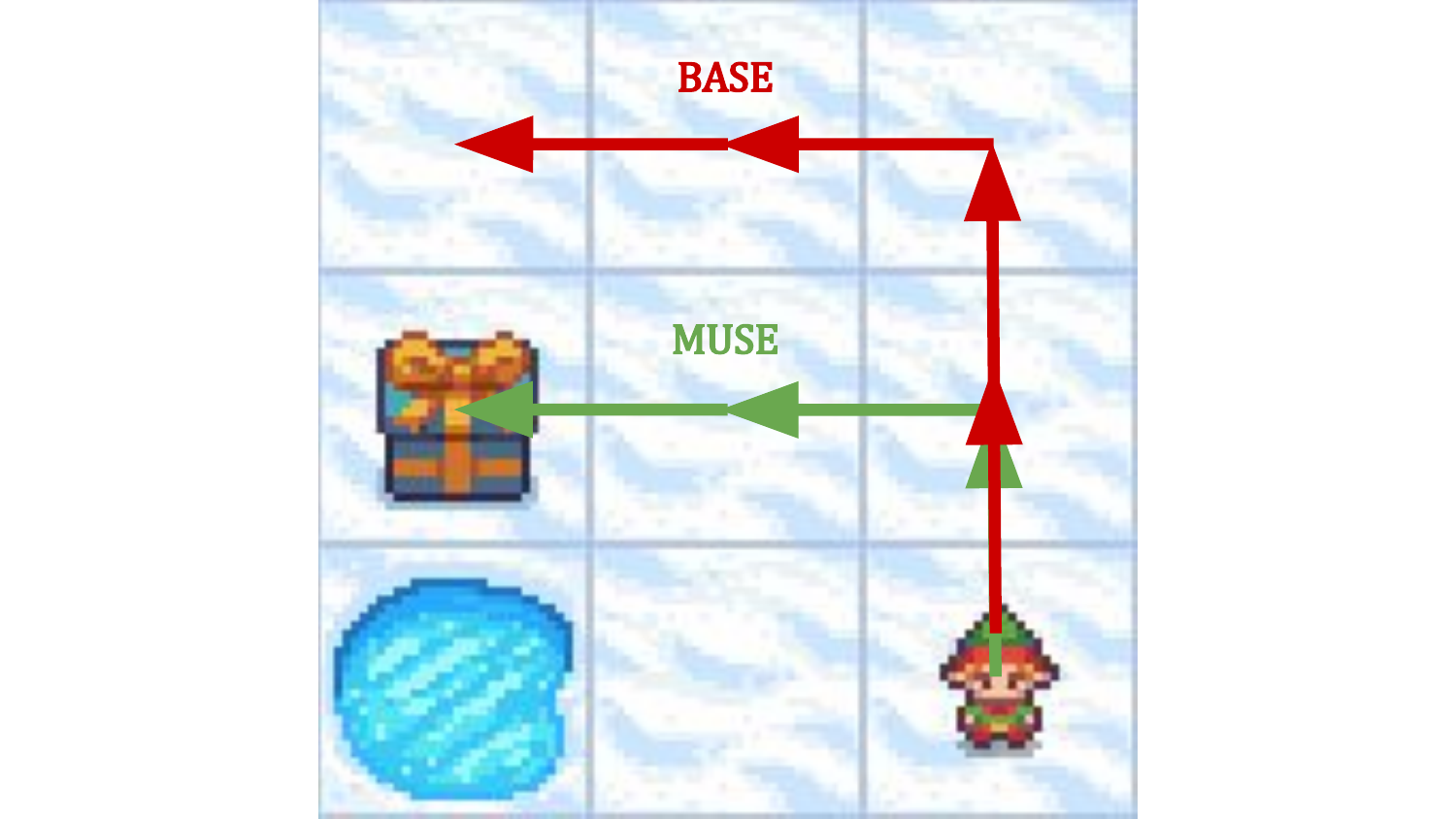}
  \caption*{(a) GPT-4o / Level 3}
  \end{subfigure}
  \begin{subfigure}{0.245\linewidth}
  \includegraphics[scale=0.275,trim=155 0 155 0, clip]{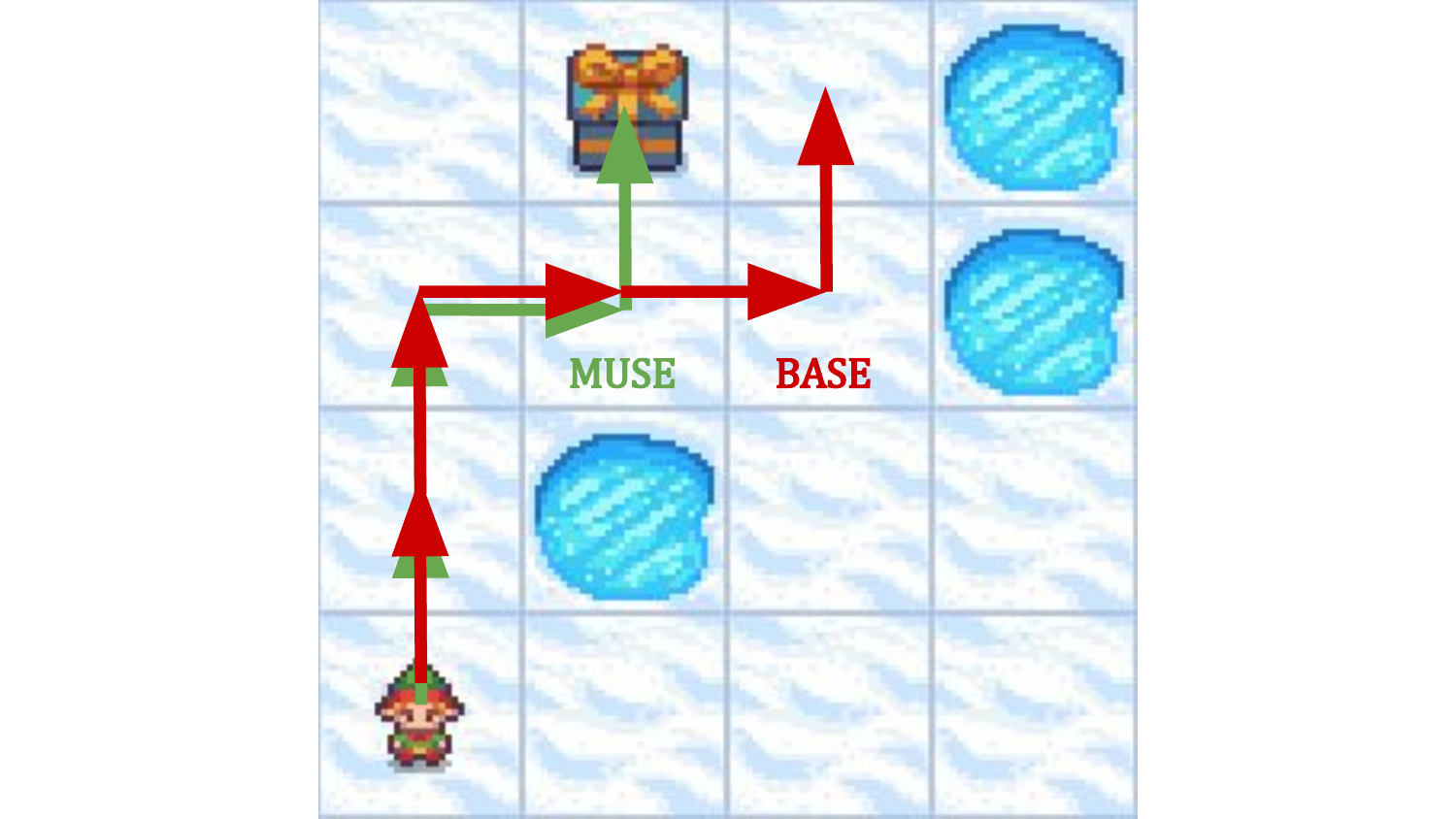}
  \caption*{(b) GPT-5.4 / Level 4}
  \end{subfigure}
  \begin{subfigure}{0.245\linewidth}
  \includegraphics[scale=0.275,trim=155 0 155 0, clip]{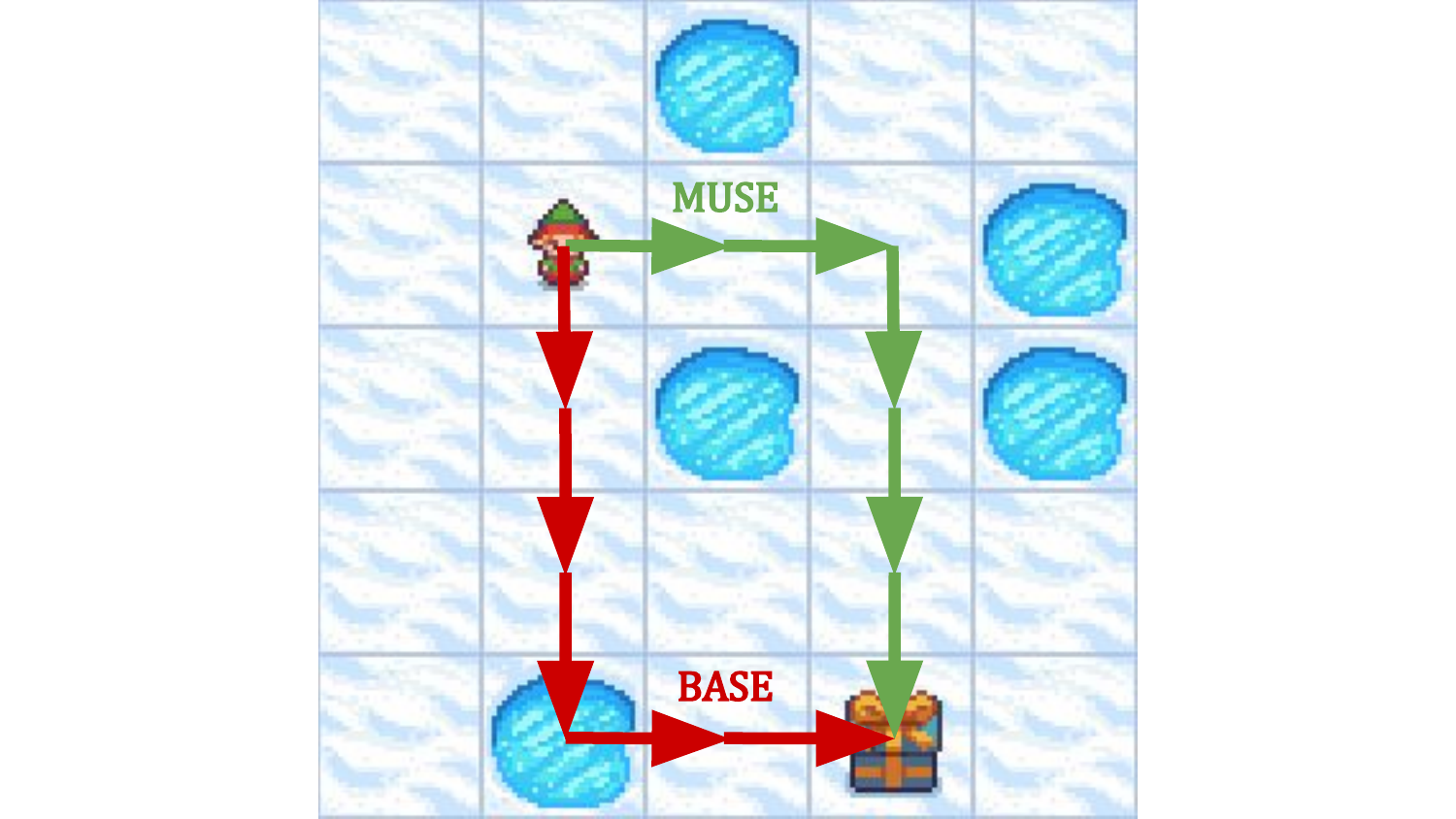}
  \caption*{(c) Haiku 4.5 / Level 5}
  \end{subfigure}
  \begin{subfigure}{0.245\linewidth}
  \includegraphics[scale=0.275,trim=155 0 155 0, clip]{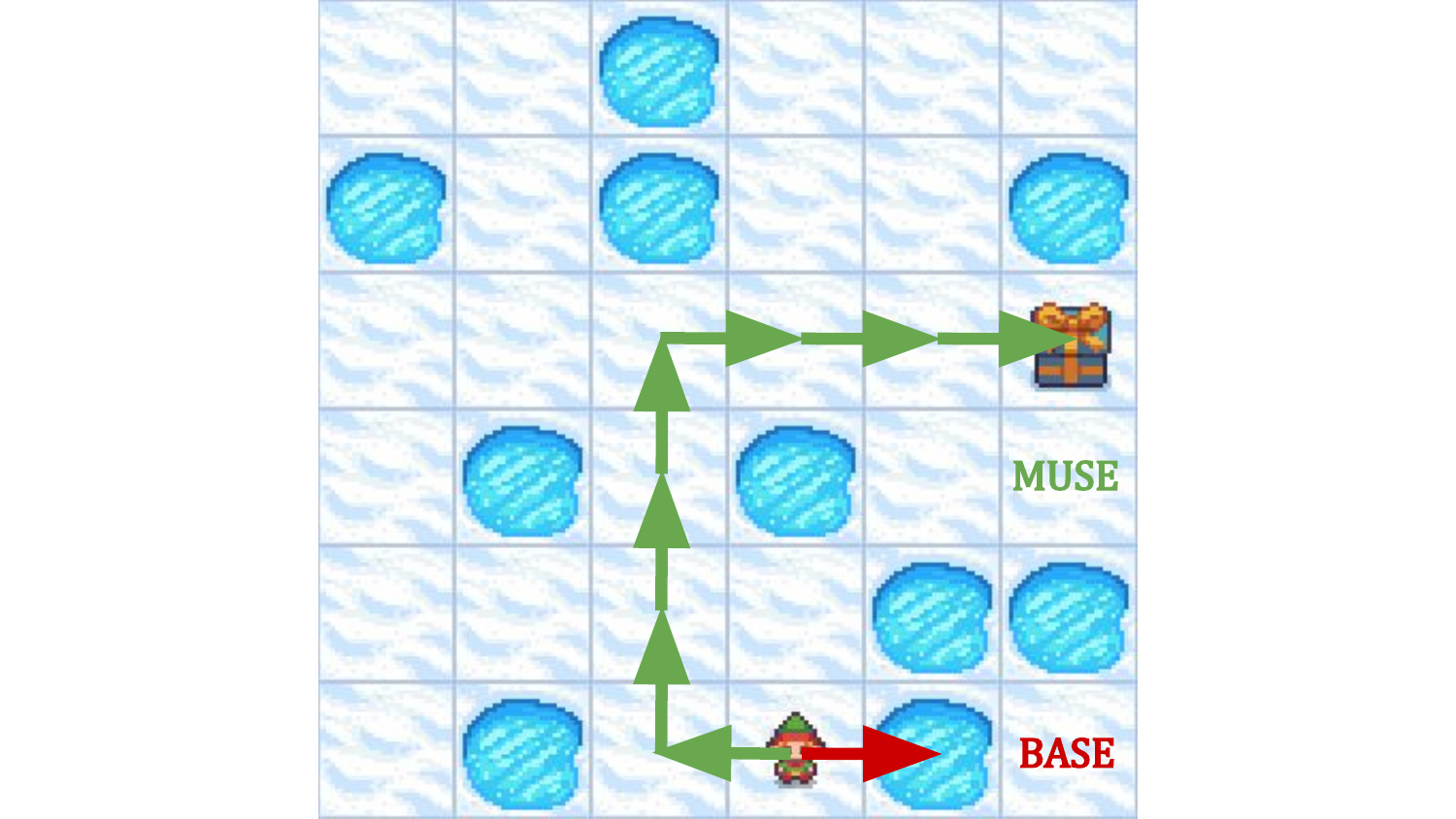}
  \caption*{(d) Opus 4.7 / Level 6}
  \end{subfigure}
  \caption{Trajectory comparison between \texttt{BASE} (red) and \texttt{MUSE} (green) on the VSP-Grid task. Each panel shows one MLLM at one difficulty level. \texttt{MUSE} produces shorter, more goal-directed paths than \texttt{BASE}.}
  \label{figure_vsp}
\end{figure*}

\paragraph{Models.}
We evaluate MUSE using four frozen MLLMs: OpenAI GPT-4o~\citep{openai2024gpt4o}, OpenAI GPT-5.4~\citep{singh2026openaigpt5card}, Anthropic Claude Haiku~4.5~\citep{claudehaiku}, and Anthropic Claude Opus~4.7~\citep{claudeopus}.
We compare two execution modes under a controlled setting that shares identical testing sets:
i) \texttt{BASE}, which performs a single MLLM call followed by deterministic parsing and verification, serving as the zero-shot baseline used in prior benchmarks;
ii) \texttt{MUSE}, our full agentic harness (Algorithm~\ref{alg:harness}), which uses the same initial prompt as \texttt{BASE} but enables verifier-guided repair with up to $K$ additional iterations when the outcome policy rejects the structured output.

\paragraph{Metrics.}
For VSP-Grid, we evaluate performance using two metrics: Valid Path, defined as the fraction of instances where the simulated trajectory reaches the goal without traversing hazardous cells, and Optimal Length, defined as the fraction of instances whose trajectory reaches the goal via a shortest path on the parsed grid.
For BLINK-Jigsaw, we report selection accuracy against the dataset-provided ground-truth label.
For CoMT, we report accuracy based on canonicalized answer matching.
For Word Search, we report exact-match accuracy against the dataset-provided answer. A single shared verifier handles both the counting and localization sub-types, judging a prediction correct only when it matches the reference in both type and value.
In addition, we report the total number of MLLM calls, and for \texttt{MUSE}, the number of instances that trigger at least one repair turn.

\begin{table}[!tb]
\small
\centering
\begin{minipage}{0.49\linewidth}
\vspace{1mm}
\small
\begin{questionbox2}
\par\smallskip
$AB$ is the diameter of $\odot O$.
Point $C$ is a point on $AB$.
Point $D$ is on $\odot O$ with $AD = AC$. Connect $DC$ and extend it to intersect $\odot O$ at point $E$. Connect $AD$, $AE$, $OE$.
If $\angle BAD = 30^{\circ}$, the degree measure of $\angle COE$ is\,(\,)? 
\par\medskip
\centering
{\includegraphics[scale=0.53,trim=0 0 0 0,clip]{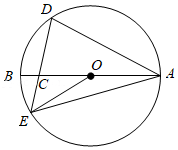}}
\par\smallskip
\textbf{Options:}\quad
\textbf{A.}~$30^{\circ}$\quad
\textbf{B.}~$35^{\circ}$\quad
\textbf{C.}~$40^{\circ}$\quad
\textbf{D.}~$45^{\circ}$ 
\begin{flushleft}
\ \ \textbf{Answer: \ \ \ A.}
\end{flushleft}
\end{questionbox2}
\end{minipage}
\hfill
\begin{minipage}{0.49\linewidth}
\centering
\caption{Results comparison on the CoMT task. ``Valid'' indicates the number of correct answers; ``Calls'' counts every model call across all problems; ``\#Repaired'' counts problems on which the harness invoked at least one repair turn. $\Delta$Valid indicates the magnitude of the improvement.}
\label{table_comt_examp}
\setlength{\tabcolsep}{1.1mm}
\begin{tabular}{lccccc}
\toprule
Model & Approach & Valid & Calls & \#Repaired & $\Delta$Valid\\
\midrule
\multirow{2}{*}{GPT-4o}              
& \texttt{BASE}  & 101 & 200 & -- & -- \\
& \texttt{MUSE} & 175 & 375 & 99 & +74 \\
\midrule
\multirow{2}{*}{GPT-5.4}             
& \texttt{BASE}  & 121 & 200 & --  & -- \\
& \texttt{MUSE} & 196 & 306 & 79 & +75 \\
\midrule
\multirow{2}{*}{Haiku 4.5}     
& \texttt{BASE}  & 144 & 200 & -- & -- \\
& \texttt{MUSE} & 182 & 303 & 51 & +38 \\
\midrule
\multirow{2}{*}{Opus 4.7}     
& \texttt{BASE}  & 161 & 200 & --  & -- \\
& \texttt{MUSE} & 197 & 249 & 40  & +{36} \\
\bottomrule
\end{tabular}
\end{minipage}
\end{table}

\subsection{MUSE for Visual Spatial Planning}
\label{sec:vsp}
Table~\ref{table1_vsp} reports the valid-path rate and optimal-length rate for all models under \texttt{BASE} and \texttt{MUSE}.
Across all models and grid sizes, \texttt{MUSE} consistently improves valid-path accuracy over \texttt{BASE}, with substantial gains at every difficulty level.
Specifically, GPT-4o improves by $+24$ points at Level~3 and $+17$ at Level~6; GPT-5.4 achieves its largest gain at Level~5 with $+31$ points; Claude~Haiku~4.5 improves by $+28$ points at Level~3.
Even for Claude~Opus~4.7, which is already strong under \texttt{BASE}, \texttt{MUSE} further increases performance to $97\%$ valid at Level~3 and $81\%$ on the hardest $6 \times 6$ grids.

The optimal-length (``O'' column) results reveal a more nuanced pattern.
For Claude~Haiku~4.5, \texttt{MUSE} achieves $70\%$ optimal at Level~3 and averages $61\%$ across all levels, indicating that most repaired trajectories are not only valid but also near-optimal once perception is corrected.
In contrast, GPT-5.4 exhibits a larger validity–optimality gap at Levels~4–6, suggesting that while the harness effectively filters invalid plans, it cannot always guide the model toward shorter trajectories within the limited repair budget $K$.

Figure~\ref{figure_vsp} provides qualitative comparisons of trajectories generated by \texttt{BASE} and \texttt{MUSE} across models and difficulty levels.
A consistent pattern emerges: \texttt{MUSE} produces more direct and goal-oriented paths, whereas \texttt{BASE} often exhibits redundant movements, detours, or oscillatory behaviors (see Appendix~\ref{appendix_vsp} for additional results).
Importantly, these gains are consistent across models, suggesting that the improvements arise from the harness design rather than model-specific factors.
Qualitatively, \texttt{BASE} errors are heterogeneous, ranging from hazard violations and goal-reaching failures to perceptual mis-localization of obstacles, whereas \texttt{MUSE} yields more reliable and efficient action sequences.

\subsection{MUSE for Multimodal Reasoning}
\label{sec:comt}
In Table~\ref{table_comt_examp}, the example on the left illustrates the CoMT task, which requires multi-step mathematical reasoning over structured multimodal inputs.
As shown in Table~\ref{table_comt_examp}, across all models, \texttt{MUSE} yields substantial improvements over \texttt{BASE}, with gains of +74 (GPT-4o), +75 (GPT-5.4), +38 (Claude~Haiku~4.5), and +36 (Claude~Opus~4.7) in valid answers.
These large margins suggest that CoMT is highly sensitive to reasoning quality, where small errors in intermediate steps can propagate to incorrect final predictions.
Detailed case studies (Appendix~\ref{appendix_comt_examples}) reveal consistent failure patterns under \texttt{BASE}, including incomplete reasoning chains, inconsistencies between intermediate reasoning and final answers, and parsing failures (e.g., malformed outputs despite correct reasoning).
\texttt{MUSE} addresses these failures through iterative, verifier-guided repair, indicating that many CoMT errors arise from correctable reasoning or execution issues rather than fundamental capability limitations.
Additional qualitative comparisons are provided in Appendix~\ref{appendix_comt_examples}.

\subsection{MUSE for Fine-Grained Visual Discrimination}
\label{sec:word_search}

The example on the left of Table~\ref{tab_word_search} illustrates the Word Search task, in which the model is asked to locate a designated Chinese character within a dense two-dimensional grid populated by many visually similar glyphs and respond with its $[\text{row},\text{col}]$ coordinates.
This task stresses fine-grained visual discrimination: the visual scene is highly repetitive, the target glyph occupies one cell among hundreds, and a correct answer requires per-cell attention rather than scene-level summarization.
As shown in Table~\ref{tab_word_search}, every frozen MLLM struggles under \texttt{BASE}, confirming that dense-grid discrimination is a regime where current MLLMs are perceptually unreliable.
For this task, the harness registers OCR as a perception tool (Section~\ref{sec:framework:components}) and folds its output into the initial prompt.
With this OCR evidence and verifier-guided repair, \texttt{MUSE} consistently lifts every model.
The repair counts indicate that the verifier rejects unreliable initial guesses on the majority of instances. 
The model, prompted to re-examine the image with OCR evidence in mind, often shifts to a more careful localization or recount.
Relative gains are especially large for GPT-4o and Claude Haiku~4.5, both of which improve their valid counts by approximately a factor of seven.
This pattern suggests that the bottleneck on Word Search is genuine perceptual difficulty rather than format errors or short-form parsing, and that combining a perception tool with verifier-guided repair adds value even when the underlying MLLM has very limited initial competence on a task.
Additional qualitative comparisons are provided in Appendix~\ref{appendix_word_search_examples}.

\begin{table}[!tb]
\small
\centering
\begin{minipage}{0.52\linewidth}
\vspace{1mm}
\small
\begin{questionbox4}
\par\smallskip
\begin{CJK}{UTF8}{gbsn}
请判断图片中的“话”字位于第几行、第几列。最后输出两个数字：
第一个数字表示从上到下的行号；
第二个数字表示从左到右的列号。
例如 [1, 2] 表示第一行第二列。
\end{CJK}
\par\medskip
\centering
{\includegraphics[scale=0.09,trim=6 4 9 15,clip]{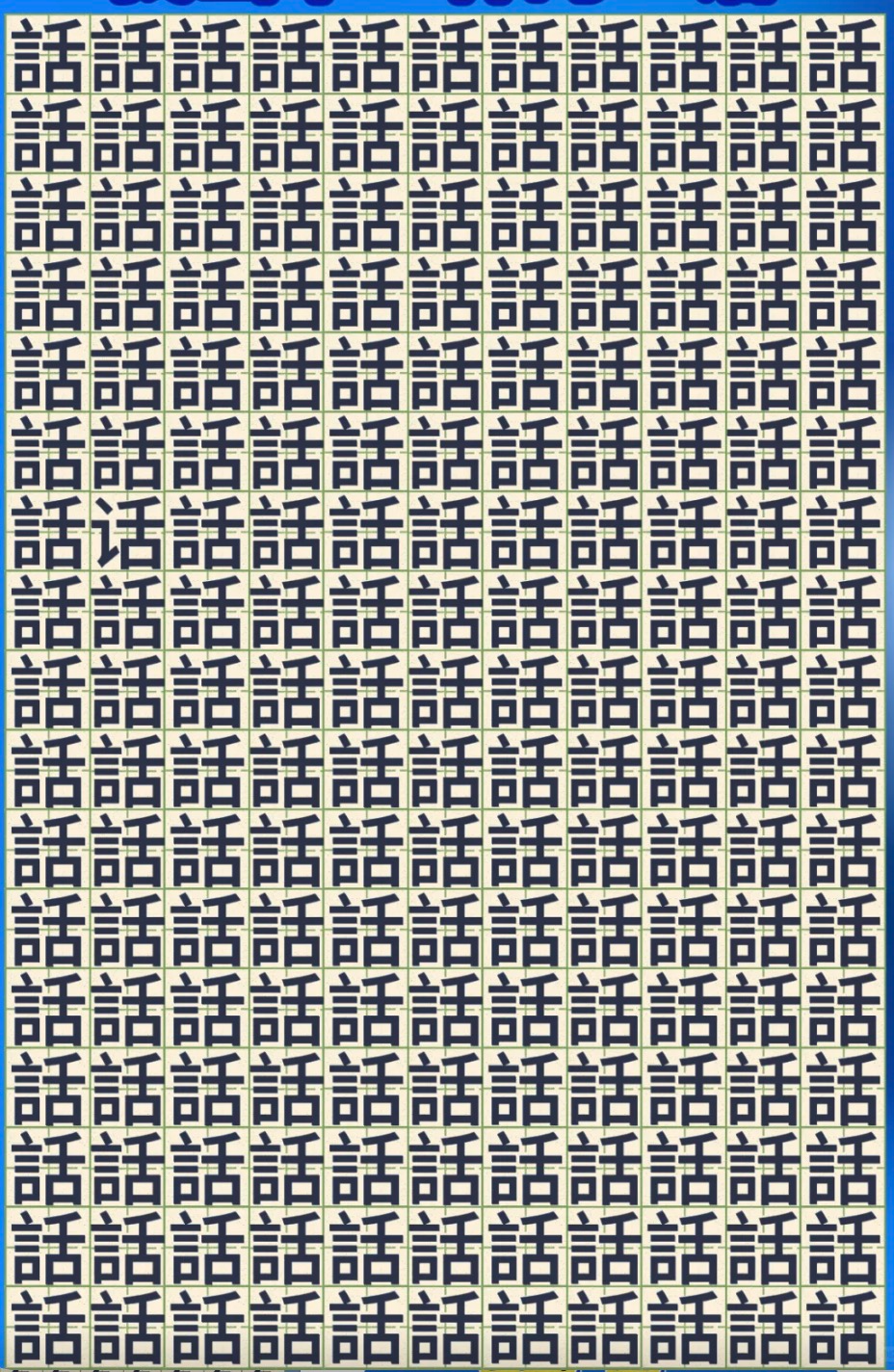}}
\begin{flushleft}
\textbf{Answer: \ [7, 2].}
\end{flushleft}
\end{questionbox4}
\end{minipage}
\hfill
\begin{minipage}{0.46\linewidth}
\centering
\caption{Results comparison on the Word Search task. ``Valid'' indicates the number of correct answers; ``Calls'' counts every model call across all problems; ``\#Repaired'' counts problems on which the harness invoked at least one repair turn. $\Delta$Valid indicates the magnitude of the improvement.}
\label{tab_word_search}
\setlength{\tabcolsep}{0.7mm}
\begin{tabular}{lccccc}
\toprule
Model & Approach & Valid & Calls & \#Repaired & $\Delta$Valid\\
\midrule
\multirow{2}{*}{GPT-4o}              
& \texttt{BASE}  & 3 & 100 & -- & -- \\
& \texttt{MUSE} & 21 & 352 & 88 & +18  \\
\midrule
\multirow{2}{*}{GPT-5.4}
& \texttt{BASE}  & 30 & 100 & --  & -- \\
& \texttt{MUSE} & 62 & 246 & 57 & +32  \\
\midrule
\multirow{2}{*}{Haiku 4.5}     
& \texttt{BASE}  & 5 & 100 & -- & -- \\
& \texttt{MUSE}  & 35 & 315 & 78 & +30   \\
\midrule
\multirow{2}{*}{Opus 4.7}     
& \texttt{BASE}  & 23 & 100 & --  & -- \\
& \texttt{MUSE} & 59 & 251 & 59 & +36   \\
\bottomrule
\end{tabular}
\end{minipage}
\end{table}

\subsection{MUSE for Visual Perception}
\label{sec:blink}
The example on the left of Table~\ref{tab111_blink} illustrates the BLINK-Jigsaw task, where the objective is to identify the correct patch to complete an incomplete image based on visual context.
As shown in Table~\ref{tab111_blink}, \texttt{MUSE} consistently improves over \texttt{BASE} across all evaluated models, indicating that many initial prediction errors are correctable through verifier-guided repair. 
Qualitative inspection reveals a common failure mode: \texttt{BASE} frequently selects patches with poor seam continuity despite plausible semantic content. 
Such errors are effectively detected by the seam-stitch verifier via pixel-level $L_1$ mismatch along patch boundaries, providing a reliable signal for repair. 
While the magnitude of the gains varies across models, the improvements are uniformly positive, highlighting the generality of the proposed harness.
Additional qualitative comparisons are provided in Appendix~\ref{appendix_BLINK_examples}.

\begin{table}[!tb]
\small
\centering
\begin{minipage}{0.49\linewidth}
\vspace{1mm}
\small
\begin{questionbox3}
\par\smallskip
Given the first image with the lower right corner missing, which one is the missing part \,(\,)?
\par\medskip
\centering
{\includegraphics[scale=0.45,trim=165 50 230 80,clip]{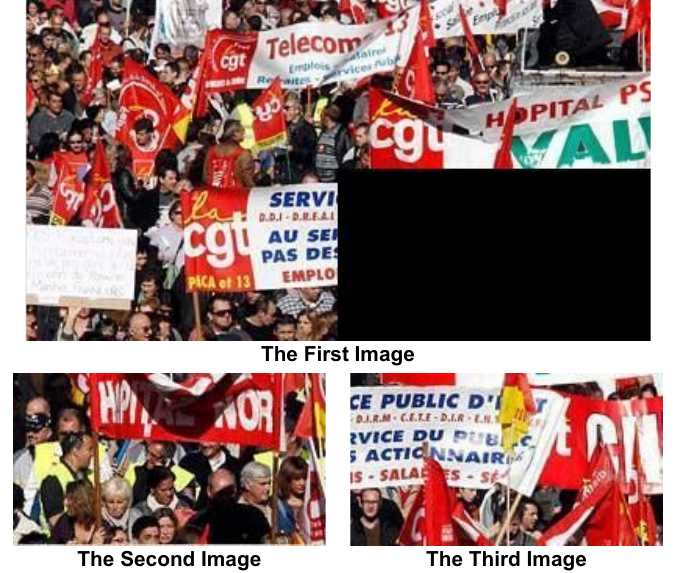}}
\par\smallskip
\textbf{Options:}\ \ 
\textbf{A.}~the second image\ \ 
\textbf{B.}~the third image
\begin{flushleft}
\textbf{Answer: \ B.}
\end{flushleft}
\end{questionbox3}
\end{minipage}
\hfill
\begin{minipage}{0.48\linewidth}
\centering
\caption{Results comparison on the BLINK-Jigsaw task. ``Valid'' indicates the number of correct answers; ``Calls'' counts every model call across all problems; ``\#Repaired'' counts problems on which the harness invoked at least one repair turn. $\Delta$Valid indicates the magnitude of the improvement.}
\label{tab111_blink}
\setlength{\tabcolsep}{1.0mm}
\begin{tabular}{lccccc}
\toprule
Model & Approach & Valid & Calls & \#Repaired & $\Delta$Valid\\
\midrule
\multirow{2}{*}{GPT-4o}              
& \texttt{BASE}  & 112 & 150 & -- & -- \\
& \texttt{MUSE} & 121 & 249 & 38 & +9 \\
\midrule
\multirow{2}{*}{GPT-5.4}
& \texttt{BASE}  & 118 & 150 & --  & -- \\
& \texttt{MUSE} & 144 & 217 & 35 & +26 \\
\midrule
\multirow{2}{*}{Haiku 4.5}     
& \texttt{BASE}  & 93 & 150 & -- & -- \\
& \texttt{MUSE} & 149 & 220 & 54 & +56 \\
\midrule
\multirow{2}{*}{Opus 4.7}     
& \texttt{BASE}  & 110 & 150 & --  & -- \\
& \texttt{MUSE} & 146 & 224 & 40  & +{36} \\
\bottomrule
\end{tabular}
\end{minipage}
\end{table}

\subsection{Ablation Study}
\label{sec:ablation}

\begin{figure*}[!tb]
  \centering
  \begin{subfigure}{0.329\linewidth}
  \includegraphics[scale=0.28,trim=0 0 0 0,clip]{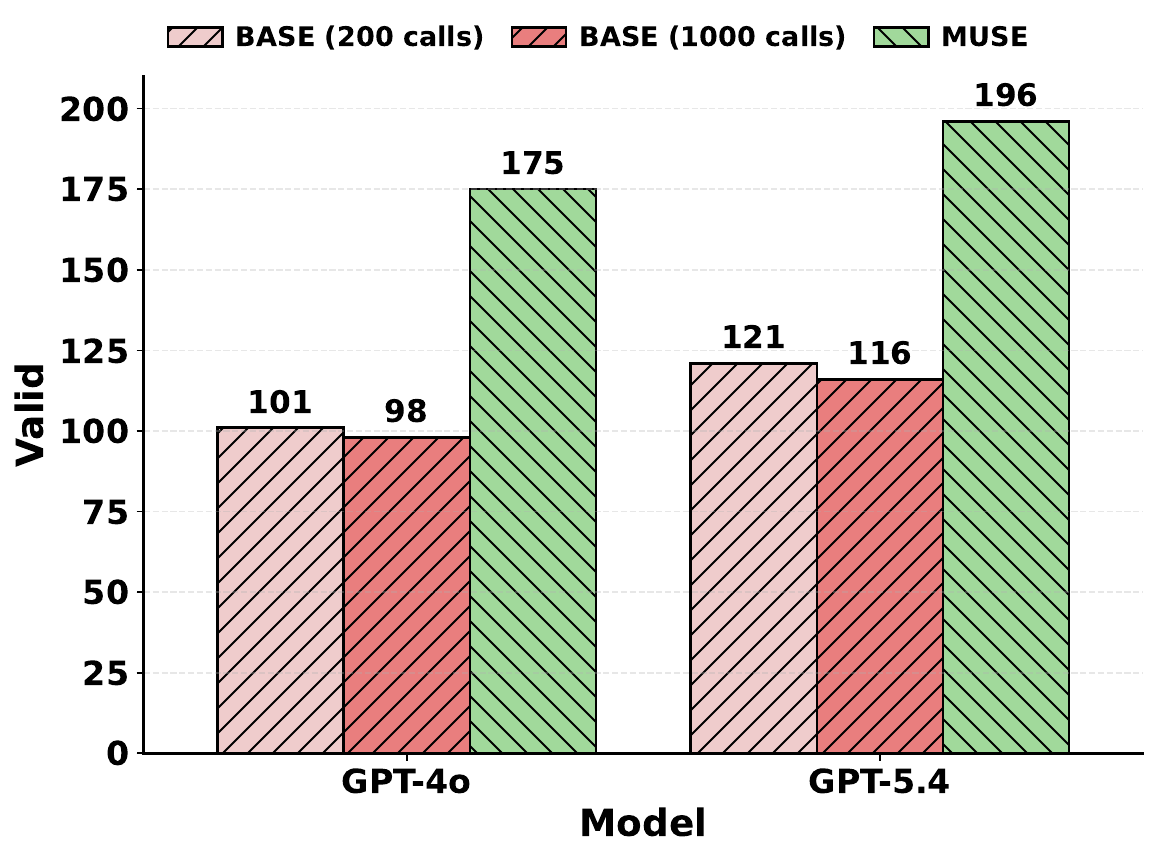}
  \caption*{(a) CoMT task}
  \end{subfigure}
  \begin{subfigure}{0.329\linewidth}
  \includegraphics[scale=0.28,trim=0 0 0 0,clip]{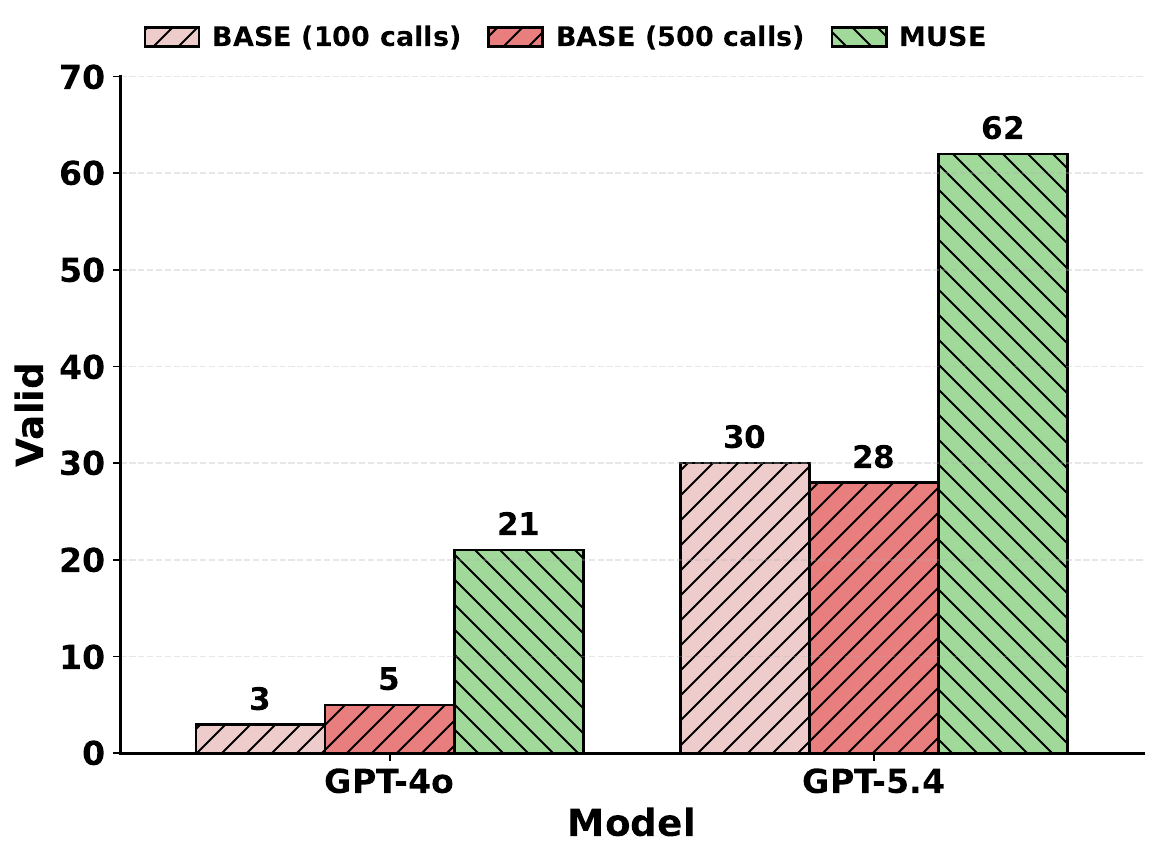}
  \caption*{(b) Word Search task}
  \end{subfigure}
  \begin{subfigure}{0.329\linewidth}
  \includegraphics[scale=0.28,trim=0 0 0 0,clip]{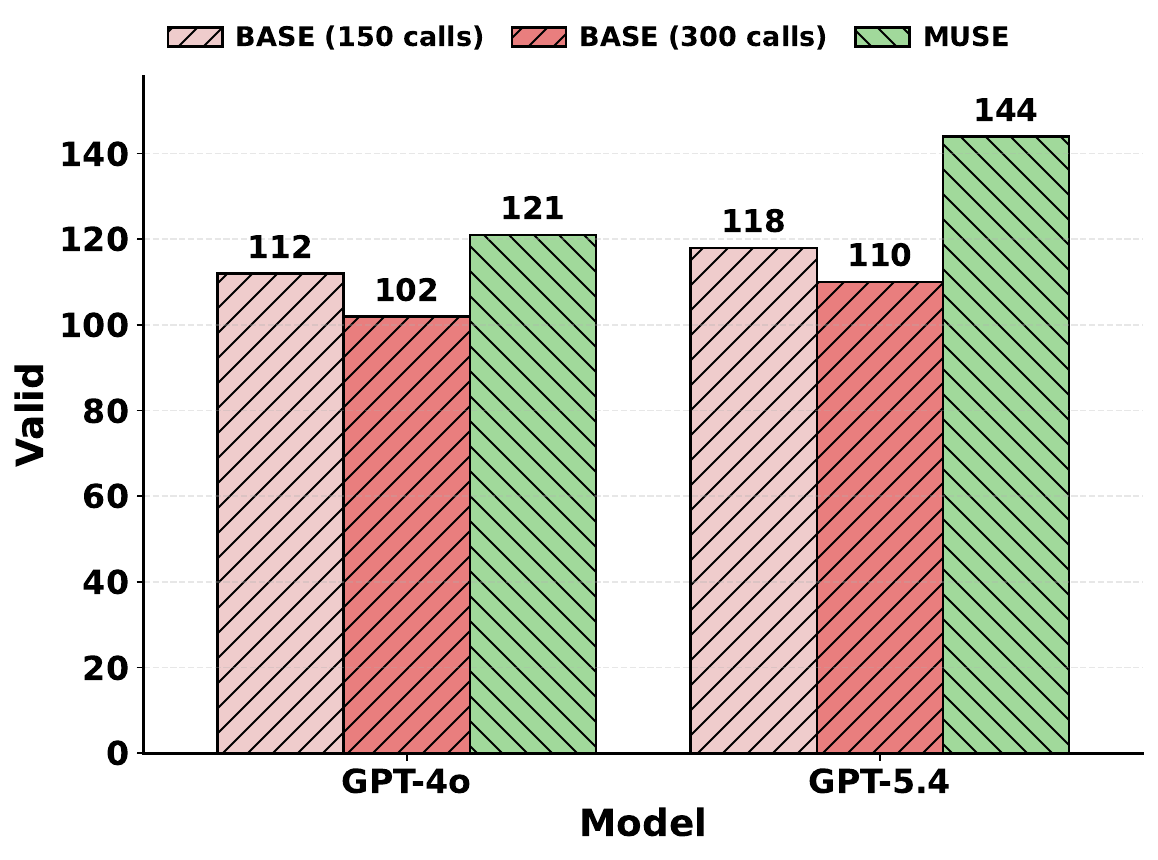}
  \caption*{(c) BLINK-Jigsaw task}
  \end{subfigure}
  \caption{Compute-matched control across three benchmarks. Results on GPT-4o and GPT-5.4 comparing the original \texttt{BASE} configuration (light red), a compute-matched self-consistency variant with additional model calls (dark red), and \texttt{MUSE} (green).  Bars denote the number of correct answers. Merely increasing the number of model calls provides little benefit and can even reduce performance. In contrast, \texttt{MUSE} consistently delivers substantial gains while using fewer model calls than the self-consistency baseline, indicating that its improvements arise from structured execution rather than a larger inference budget.}
  \label{figure_compute_matched}
\end{figure*}

\paragraph{Compute-matched control.}
One might argue that the gains of \texttt{MUSE} arise simply from issuing more model calls than \texttt{BASE}, rather than from its structured execution pipeline. 
To rule out this explanation, we grant \texttt{BASE} a call budget matched to that of \texttt{MUSE} via a self-consistency baseline that expends the additional calls without any harness structure.
Specifically, we query the model multiple times independently (five times for CoMT and Word Search, and twice for BLINK-Jigsaw) and return the majority-voted answer.
As shown in Figure~\ref{figure_compute_matched}, increasing the call budget alone yields only marginal changes in performance. On CoMT, validity changes from $101$ to $98$ for GPT-4o and from $121$ to $116$ for GPT-5.4. On Word Search, GPT-4o improves only slightly (from $3$ to $5$), while GPT-5.4 decreases from $30$ to $28$. 
On BLINK-Jigsaw, performance also changes little, dropping from $112$ to $102$ for GPT-4o and from $118$ to $110$ for GPT-5.4. 
Overall, simply increasing the number of model calls provides little benefit and does not reliably improve task performance.

In contrast, \texttt{MUSE} consistently delivers substantial gains across benchmarks and models, improving validity from $101$ to $175$ and from $121$ to $196$ on CoMT, from $3$ to $21$ and from $30$ to $62$ on Word Search, and from $112$ to $121$ and from $118$ to $144$ on BLINK-Jigsaw. Notably, these gains are achieved with fewer model calls than the compute-matched self-consistency baseline.
These results suggest that the benefits of \texttt{MUSE} do not stem from a larger inference budget. Rather, they arise from the structured design of the harness, which converts additional computation into meaningful improvements in decision quality.

\paragraph{Effect of repair budget $K$.}
\begin{wrapfigure}{r}{0.45\linewidth}
    \centering
    \includegraphics[scale=0.35,trim=0 0 0 0,clip]{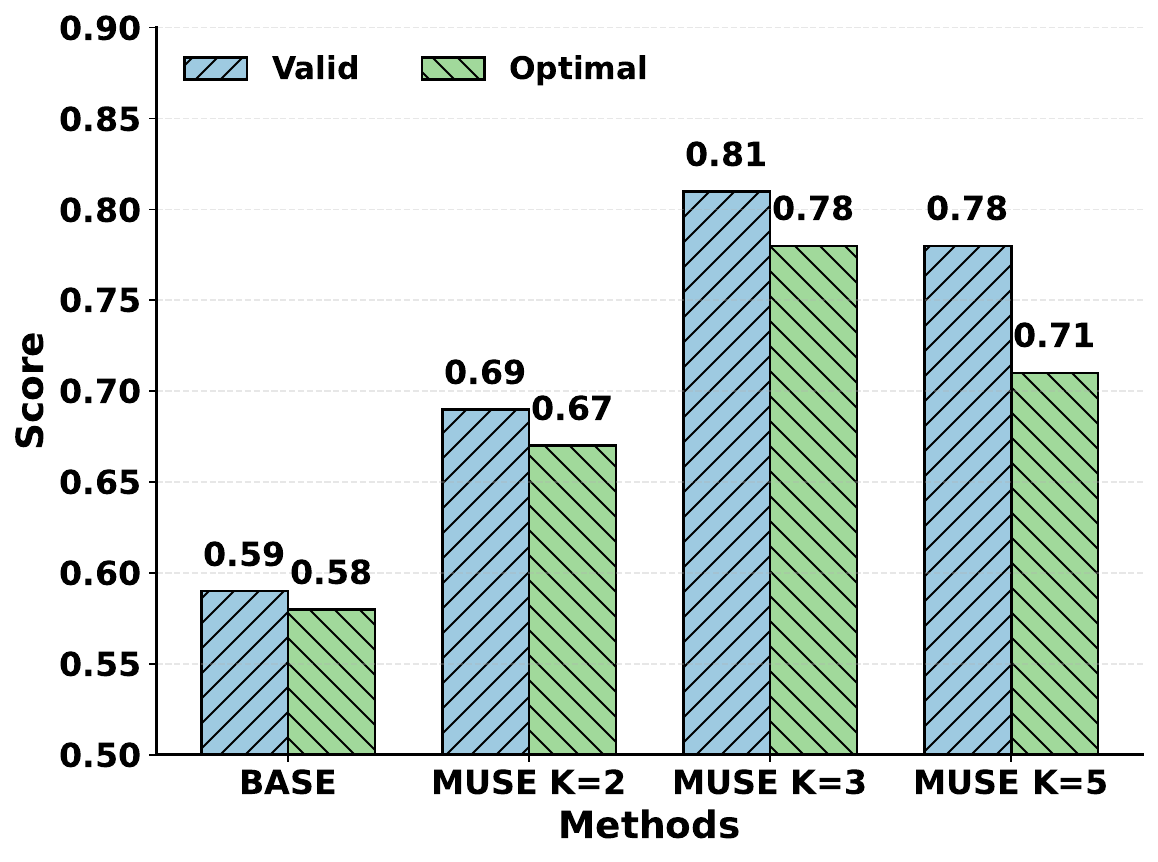}
    \captionsetup{justification=centering}
    \caption{Validity and optimality vs.\ repair budget $K$.}
    \label{budget_ablation}
\end{wrapfigure}
All headline experiments use a maximum of $K\!=\!3$ repair turns. 
However, the relationship between performance and $K$ is not necessarily monotonic.
To examine this effect, we evaluate Claude~Opus~4.7 on VSP Level~6 under varying repair budgets, $K \in \{2,3,5\}$. 
We focus on Level~6 because it is the most challenging setting and remains difficult even for the strongest model in our evaluation suite.
Figure~\ref{budget_ablation} reports both validity and optimality as functions of $K$.
Increasing the budget from $K\!=\!2$ to $K\!=\!3$ yields the largest gains, with validity improving from $0.69$ to $0.81$ and optimality from $0.67$ to $0.78$. 
This indicates that a substantial fraction of instances require at least three repair iterations to converge to correct or near-optimal solutions.
In contrast, further increasing the repair budget from $K=3$ to $K=5$ leads to lower validity ($0.78$) and optimality ($0.71$), suggesting diminishing returns and potential over-correction at larger budgets.
Notably, the drop is more pronounced for optimality ($-7$ points) than validity ($-3$ points), suggesting that additional repair turns are particularly prone to converting near-optimal trajectories into valid but suboptimal ones.
We therefore adopt $K\!=\!3$ as the default setting, as it provides a favorable trade-off between accuracy and computational cost while avoiding over-correction effects observed at larger budgets.

\paragraph{Failure-mode analysis.}
We analyze performance on the hardest grid setting ($6 \times 6$) using Claude~Opus~4.7.
Under \texttt{BASE}, the model fails on 41 instances, with the majority due to hazard violations (37 cases) and only a small fraction due to goal-reaching failures (4 cases).
Moreover, failed trajectories are typically longer than the shortest valid paths, indicating inefficient planning behavior.
Applying \texttt{MUSE} improves validity from $59\%$ to $81\%$, with repair triggered on 35 instances.
Relative to \texttt{BASE}, \texttt{MUSE} corrects 23 of the 41 initial failures, including 19 hazard violations and 4 failures to reach the goal.
This indicates that goal-reaching failures are correctable, while hazard-related errors are more persistent.
In particular, only about half of the hazard violations ($19/37$) are corrected within the repair budget $K\!=\!3$.
This suggests that, at higher difficulty levels, the dominant bottleneck shifts from high-level planning errors to low-level perceptual inaccuracies, specifically the mis-localization of hazards in the environment.

\section{Conclusion}
\label{sec:conclusion}
We presented \texttt{MUSE}, a multimodal unified structured execution harness that improves the capabilities of frozen MLLMs by optimizing the execution scaffold rather than the model itself. 
By decomposing inference into modular components, MUSE transforms single-pass generation into a structured and iterative decision-making process tailored to visual-centric tasks. Across a diverse set of benchmarks spanning visual spatial planning, visual perception, multimodal reasoning, and fine-grained visual discrimination, MUSE consistently yields substantial performance gains.
Our analysis shows that many failures stem from harness-level shortcomings rather than fundamental model deficits, establishing the multimodal harness as an optimization dimension complementary to model-centric scaling.
As frontier MLLMs continue to scale and retraining becomes increasingly expensive, harness engineering offers a practical avenue for improving system performance.
More broadly, MUSE advocates a shift in perspective: rather than focusing exclusively on building stronger models, we should also explore how to better structure, verify, and leverage the capabilities of existing ones. We hope this work encourages further research on multimodal harness design as a complementary pathway toward more capable, reliable, and efficient multimodal systems.

\textbf{Limitations}. 
Despite its effectiveness, the proposed MUSE has several potential limitations.
First, while the harness can correct many errors, it cannot fully resolve persistent perceptual failures (e.g., hazard mis-localization in complex scenes), which ultimately require stronger underlying visual representations.
Second, MUSE depends on the availability and quality of task-specific verifiers, so it is harder to apply to open-ended generation tasks, where correctness is inherently ambiguous and no reliable verification signal exists to distinguish good outputs from poor ones.
Third, improvements come at the cost of additional MLLM calls, requiring a balance between solution quality and computational cost.
Finally, our current design relies on manually constructed modules and policies. Automating harness design and adaptation is a promising avenue for future research.

\clearpage
\newpage
\bibliographystyle{plain}
\bibliography{references}

\clearpage

\appendix

\section*{Appendix}
\section{Experimental Details}
\subsection{Benchmarks}
The used benchmarks VSP-Grid, BLINK-Jigsaw, CoMT, and TIR-Bench highlight the heterogeneity in task formulations, input modalities, and expected outputs, which cover complementary capabilities of MLLMs, such as visual spatial planning, visual perception, multimodal reasoning, and fine-grained visual discrimination.

\begin{itemize}[leftmargin=15pt]
    \item VSP-Grid is a grid-based navigation task that accepts either a single image or textual input and requires the model to produce a sequence of discrete actions (U/D/L/R). This task evaluates the model's ability to plan and execute step-by-step decisions in a spatial environment. 

    \item BLINK-Jigsaw operates on multi-image inputs, where the model must select the candidate patch that completes the missing region and output a categorical answer (A or B) in a structured JSON field. This setting emphasizes visual matching and contextual reasoning across multiple images.

     \item From CoMT, we use the creation split, which pairs one or more images with a multiple-choice question and requires the model to output a single option label. The problems focus on structured multimodal reasoning, predominantly plane-geometry inference over visual inputs.
    
    \item We evaluate the Word Search split of the TIR-Bench, which operates on a single image of a dense two-dimensional grid populated with many visually similar glyphs. Each instance is one of two sub-types: a counting query whose answer is a non-negative integer giving the number of occurrences of a designated target glyph, or a localization query whose answer is an ordered integer pair $[\text{row},\text{col}]$ identifying the target's position.

\end{itemize}

Importantly, these benchmarks differ not only in modality (single image, multiple images, or text) but also in output formats, spanning move sequences, categorical letters, option labels, and either scalar integers or ordered integer pairs.
To accommodate this diversity, our proposed MUSE framework employs task-specific parsers and outcome policies, enabling consistent evaluation across heterogeneous settings while preserving the unique requirements of each task.

\subsection{Outcome Policies}
\label{appendx_outcom_police}
We currently construct the following outcome policies:
\begin{itemize}[leftmargin=15pt]
    \item Navigation (VSP-Grid). The parsed action sequence is executed in the grid simulator, and the answer is accepted only if the resulting trajectory reaches the goal without entering a hazard. Rejections carry a compact certificate (e.g.\ hazard contact, or failure to reach the goal) that can be fed back as evidence in subsequent repair turns.

    \item Jigsaw seam decisions (BLINK-Jigsaw). The model's choice is checked against an independent bottom-right seam-stitch simulator and, when ground-truth annotations are available, additionally against the reference label. Relying on the seam simulator lets the policy reward genuinely correct patch placement rather than mere agreement with a discrete label.

    \item Multiple-choice answers (CoMT). The policy canonicalizes superficial formatting variations typical of conversational decoders (e.g.\ \texttt{(B)} or \texttt{Answer: B}) and then performs exact matching on the resulting option.

    \item TIR-Bench. The verifier dispatches answers to task-specific canonicalizers, including ordered integer sequences, deduplicated integer sets, case- and punctuation-insensitive free text, and scalar letter or integer matching, providing a unified interface for evaluating heterogeneous answer formats.

    \item Fallback. When no specialized policy applies, the verifier defaults to normalized string matching against the reference answer as a conservative evaluation strategy.
\end{itemize}

\begin{figure*}[!tb]
  \centering
  \begin{subfigure}{0.245\linewidth}
  \includegraphics[scale=0.275,trim=155 0 155 0,clip]{figs/vsp_level3.pdf}
  \end{subfigure}
  \begin{subfigure}{0.245\linewidth}
  \includegraphics[scale=0.275,trim=155 0 155 0, clip]{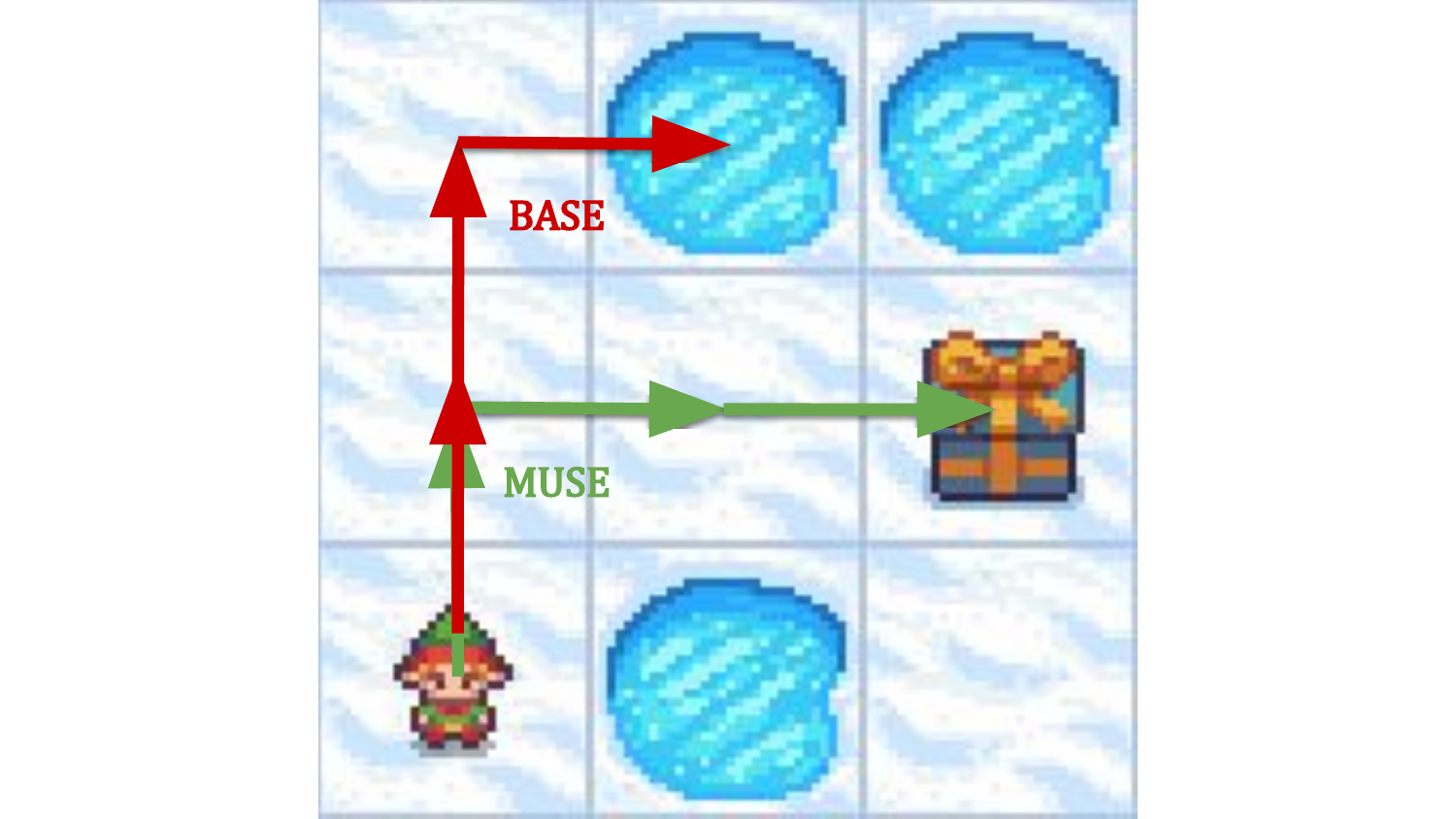}
  \end{subfigure}
  \begin{subfigure}{0.245\linewidth}
  \includegraphics[scale=0.275,trim=155 0 155 0, clip]{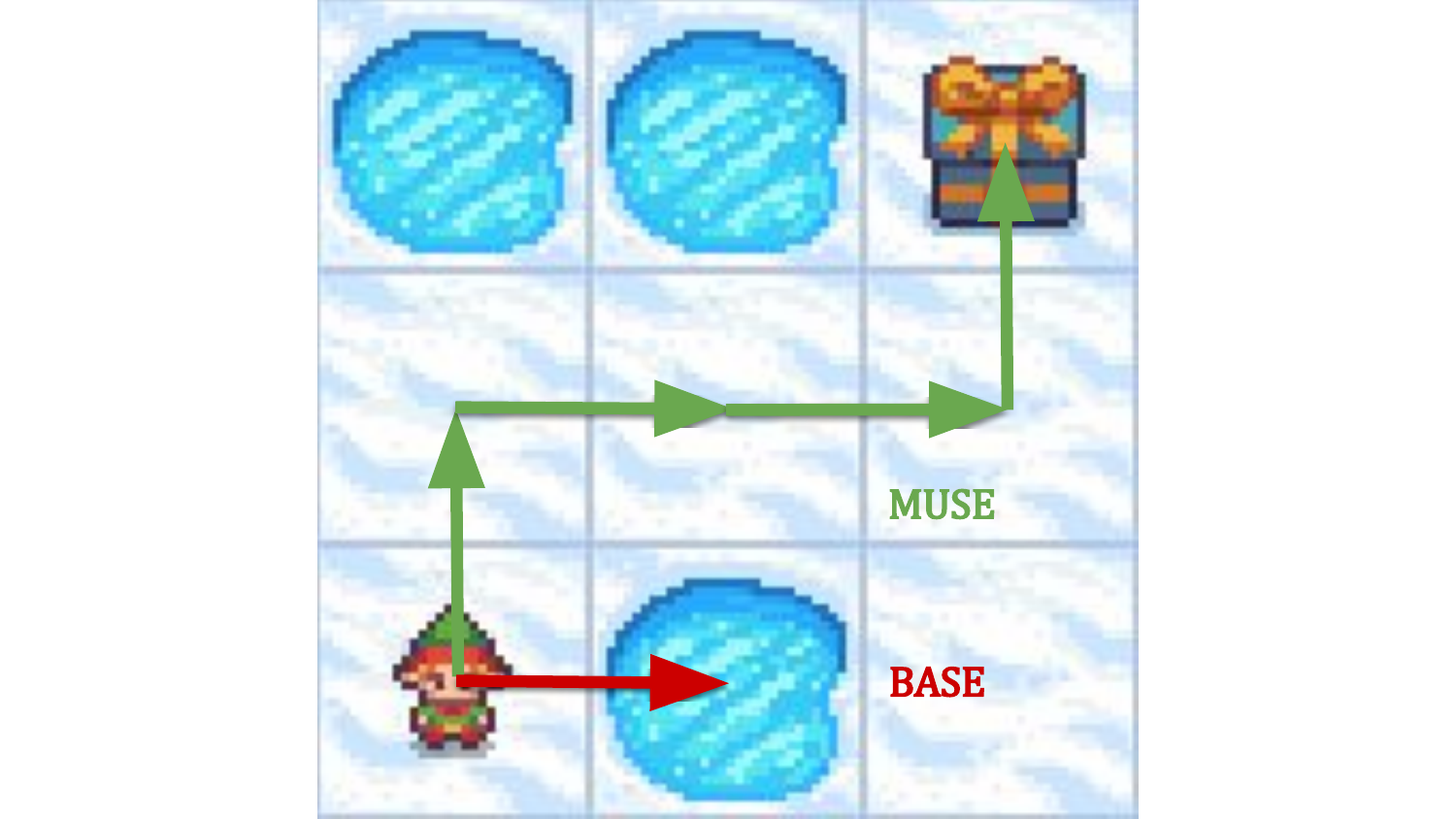}
  \end{subfigure}
  \begin{subfigure}{0.245\linewidth}
  \includegraphics[scale=0.275,trim=155 0 155 0, clip]{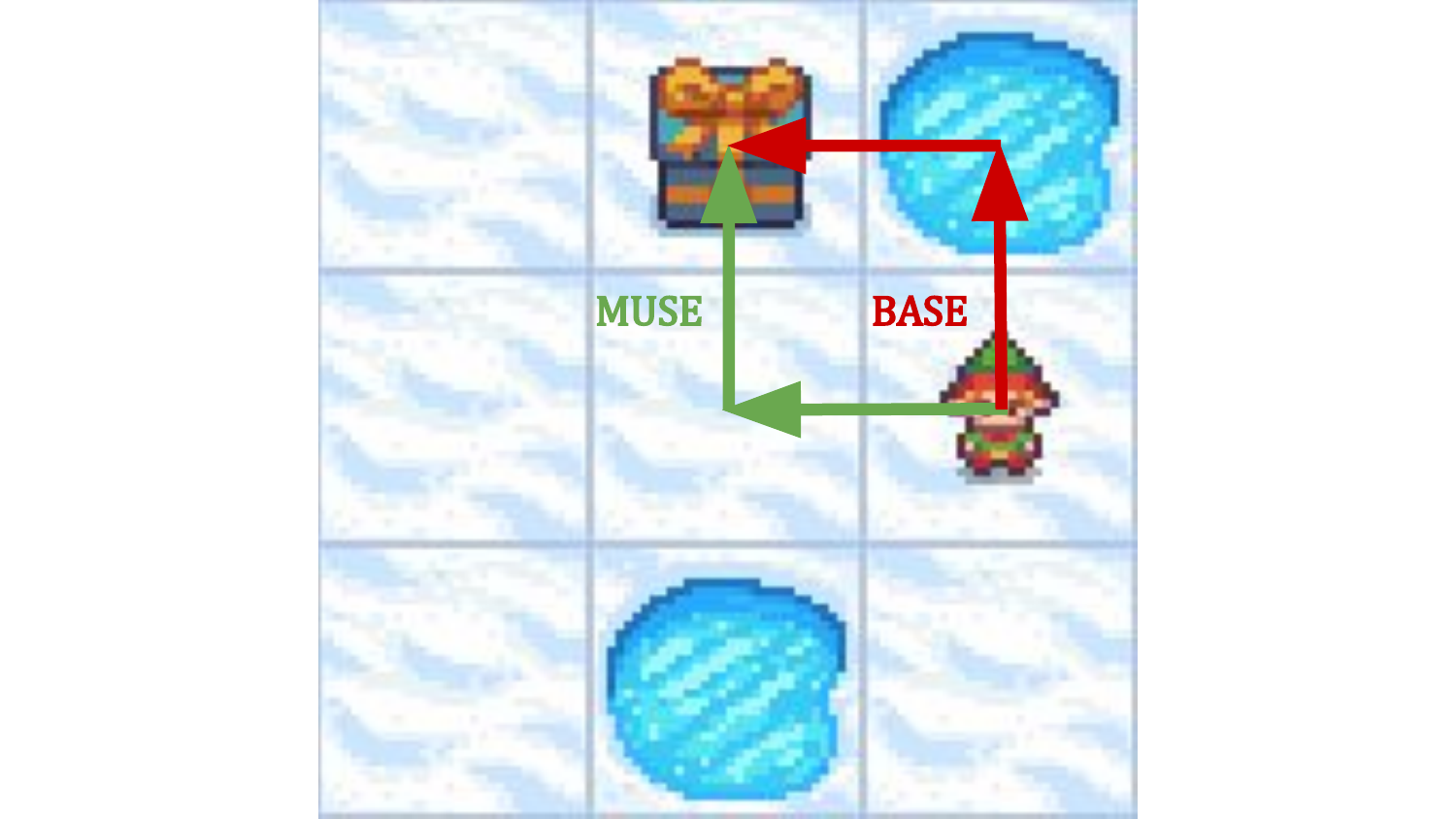}
  \end{subfigure} \\
  \vspace{1mm}
  %
  %
  \begin{subfigure}{0.245\linewidth}
  \includegraphics[scale=0.275,trim=155 0 155 0,clip]{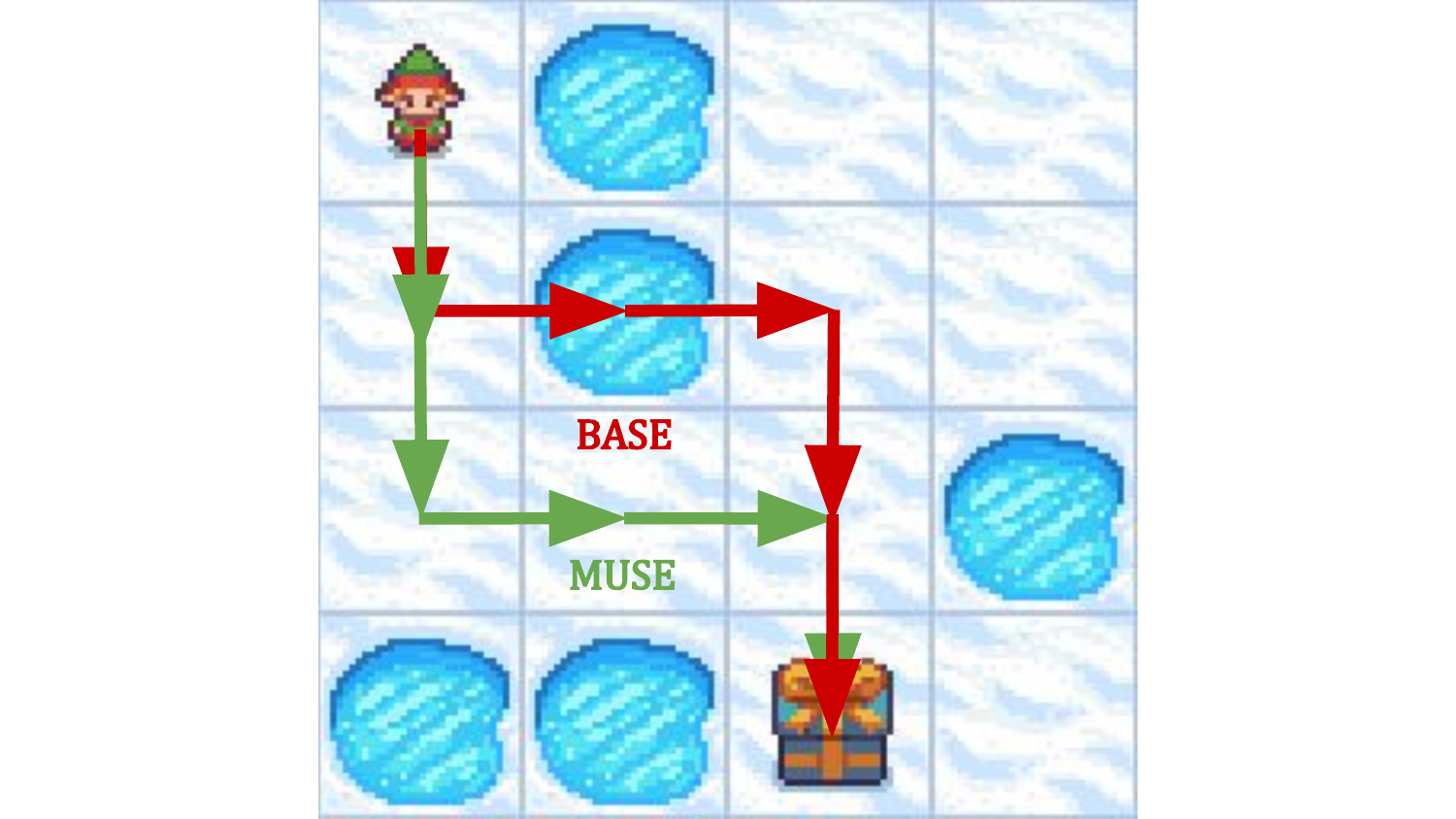}
  \end{subfigure}
  \begin{subfigure}{0.245\linewidth}
  \includegraphics[scale=0.275,trim=155 0 155 0, clip]{figs/vsp_level4.pdf}
  \end{subfigure}
  \begin{subfigure}{0.245\linewidth}
  \includegraphics[scale=0.275,trim=155 0 155 0, clip]{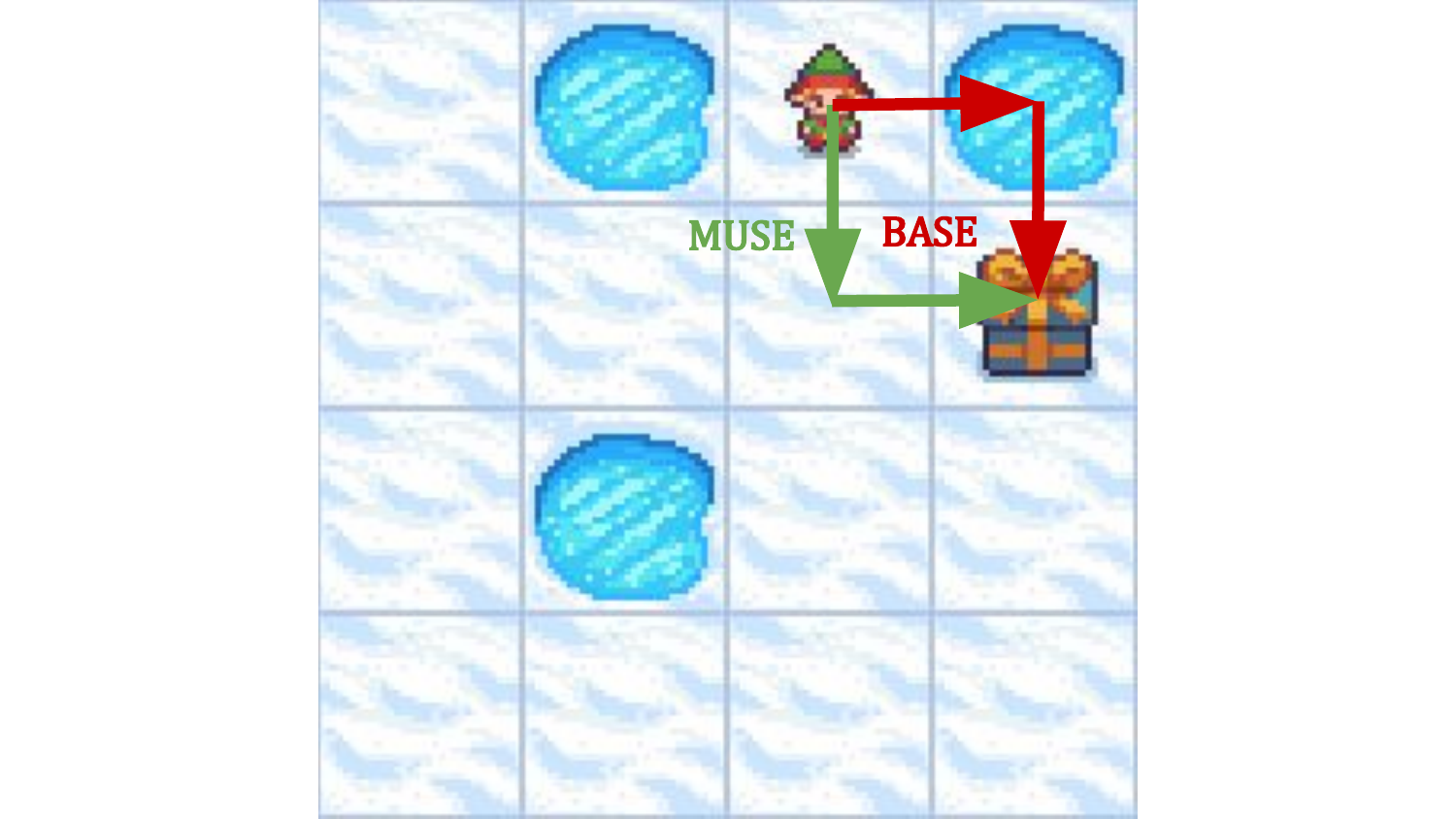}
  \end{subfigure}
  \begin{subfigure}{0.245\linewidth}
  \includegraphics[scale=0.275,trim=155 0 155 0, clip]{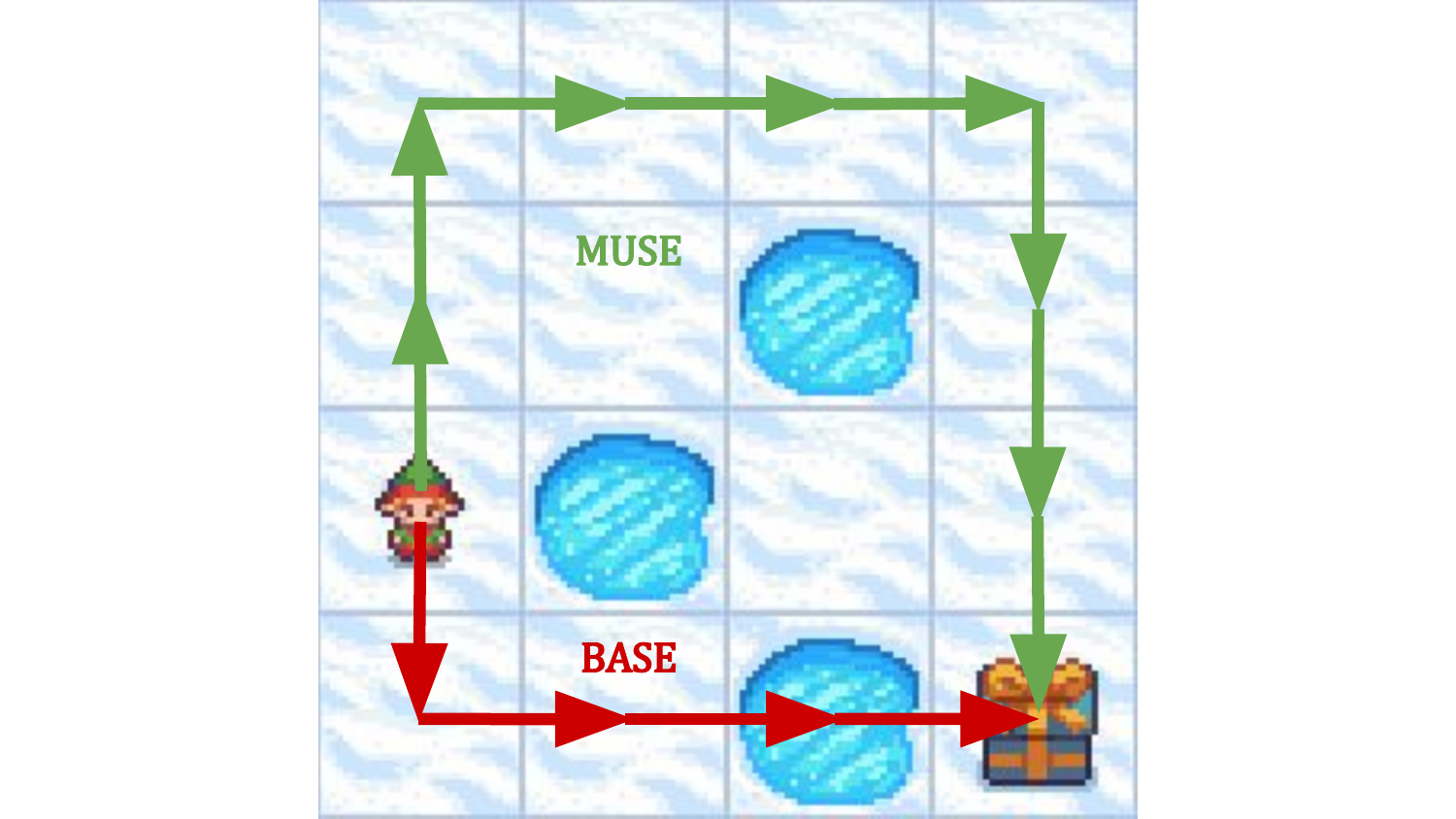}
  \end{subfigure}\\
  \vspace{1mm}
  %
  %
  \begin{subfigure}{0.245\linewidth}
  \includegraphics[scale=0.275,trim=155 0 155 0,clip]{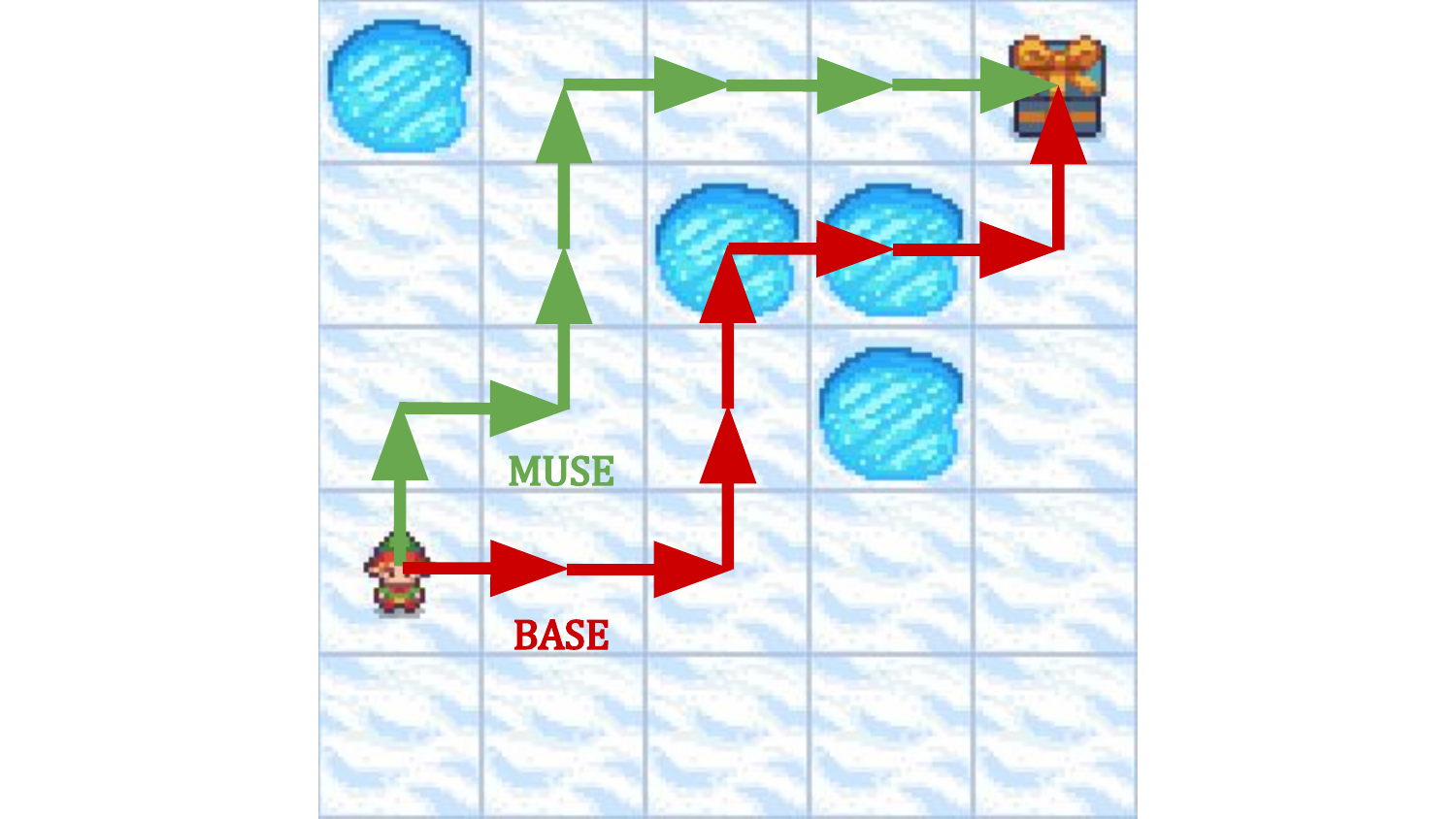}
  \end{subfigure}
  \begin{subfigure}{0.245\linewidth}
  \includegraphics[scale=0.275,trim=155 0 155 0, clip]{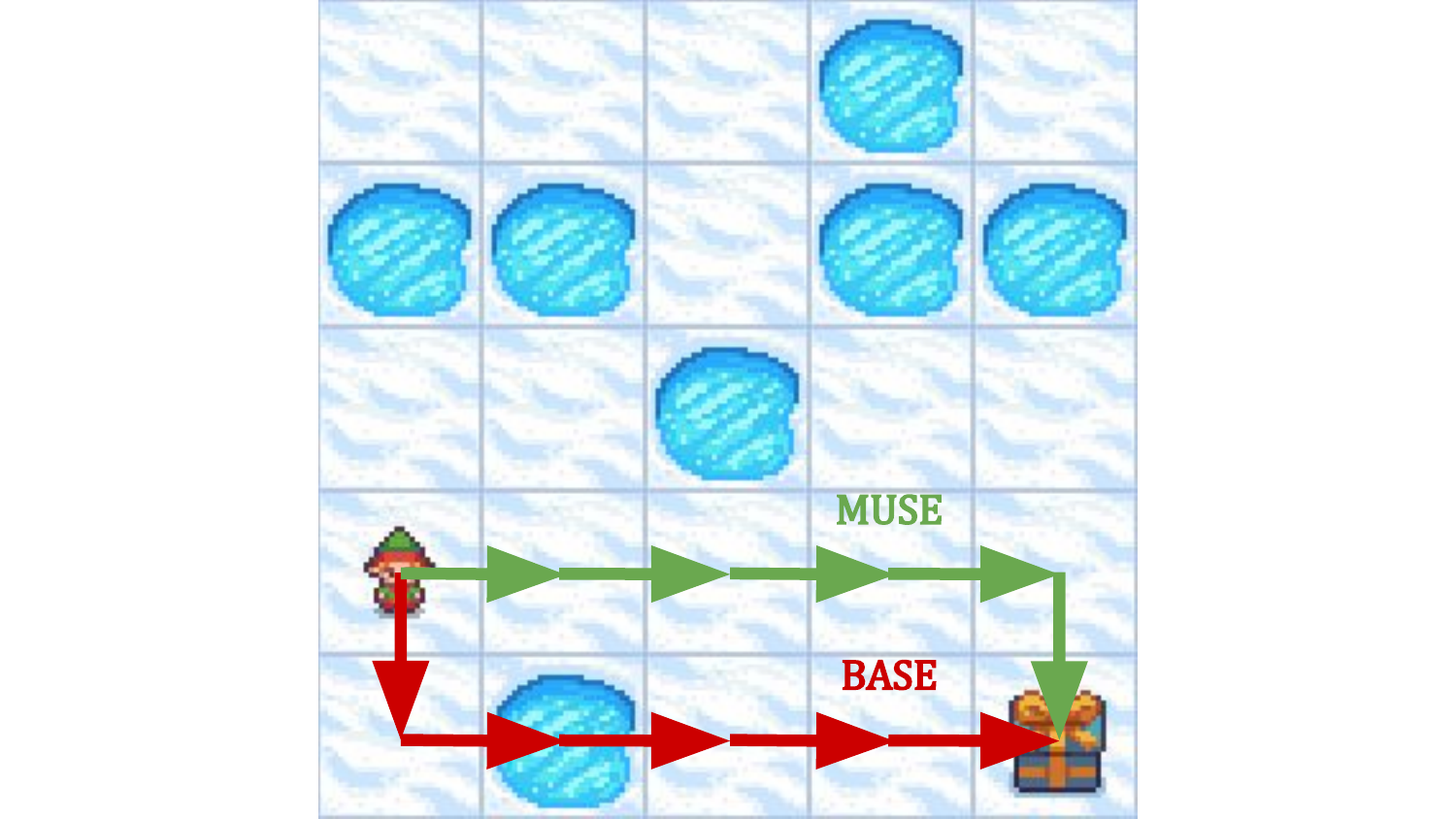}
  \end{subfigure}
  \begin{subfigure}{0.245\linewidth}
  \includegraphics[scale=0.275,trim=155 0 155 0, clip]{figs/vsp_level5.pdf}
  \end{subfigure}
  \begin{subfigure}{0.245\linewidth}
  \includegraphics[scale=0.275,trim=155 0 155 0, clip]{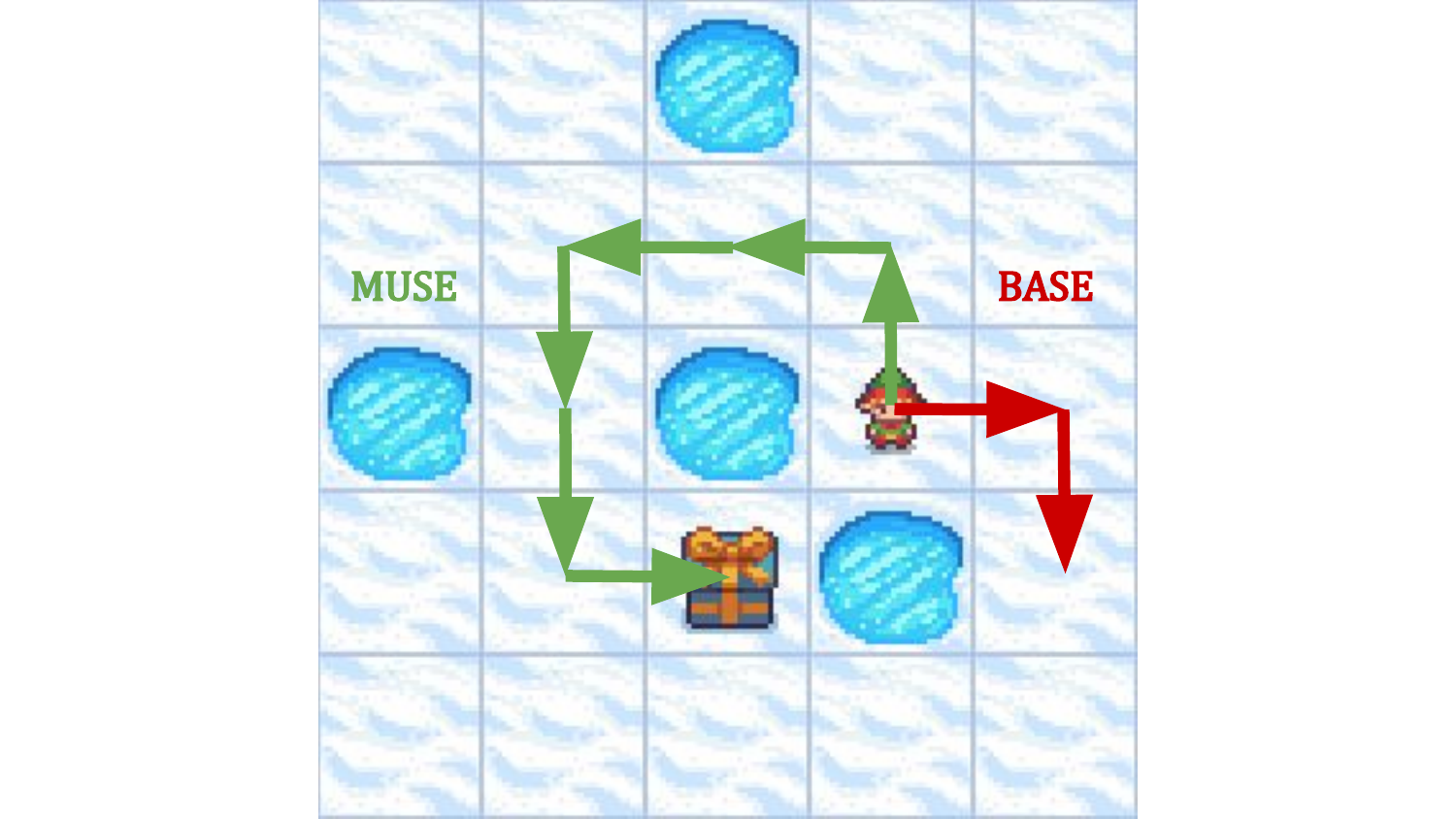}
  \end{subfigure} \\
  \vspace{1mm}
  %
  %
  \begin{subfigure}{0.245\linewidth}
  \includegraphics[scale=0.275,trim=155 0 155 0,clip]{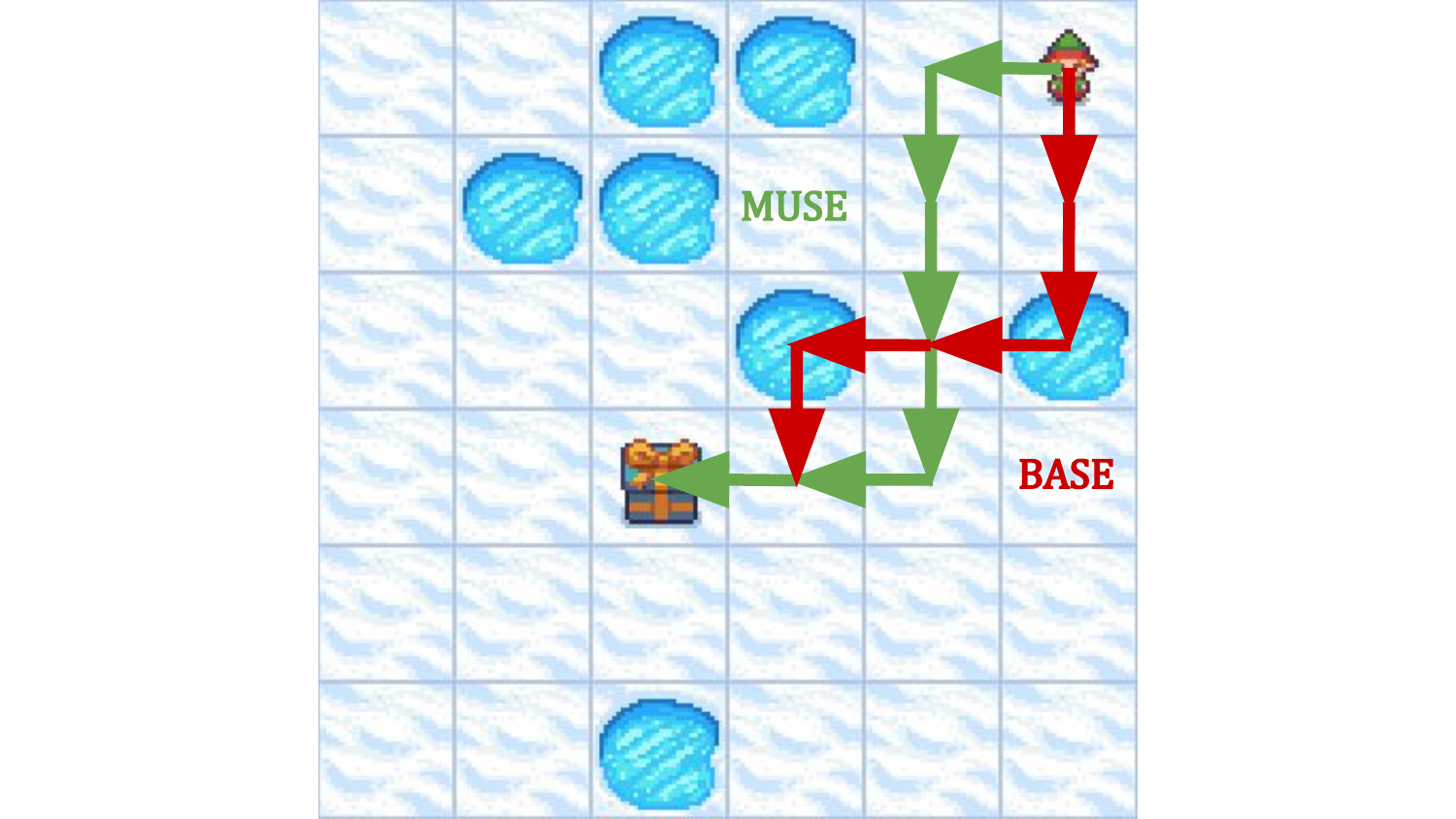}
  \caption*{(a) GPT-4o}
  \end{subfigure}
  \begin{subfigure}{0.245\linewidth}
  \includegraphics[scale=0.275,trim=155 0 155 0, clip]{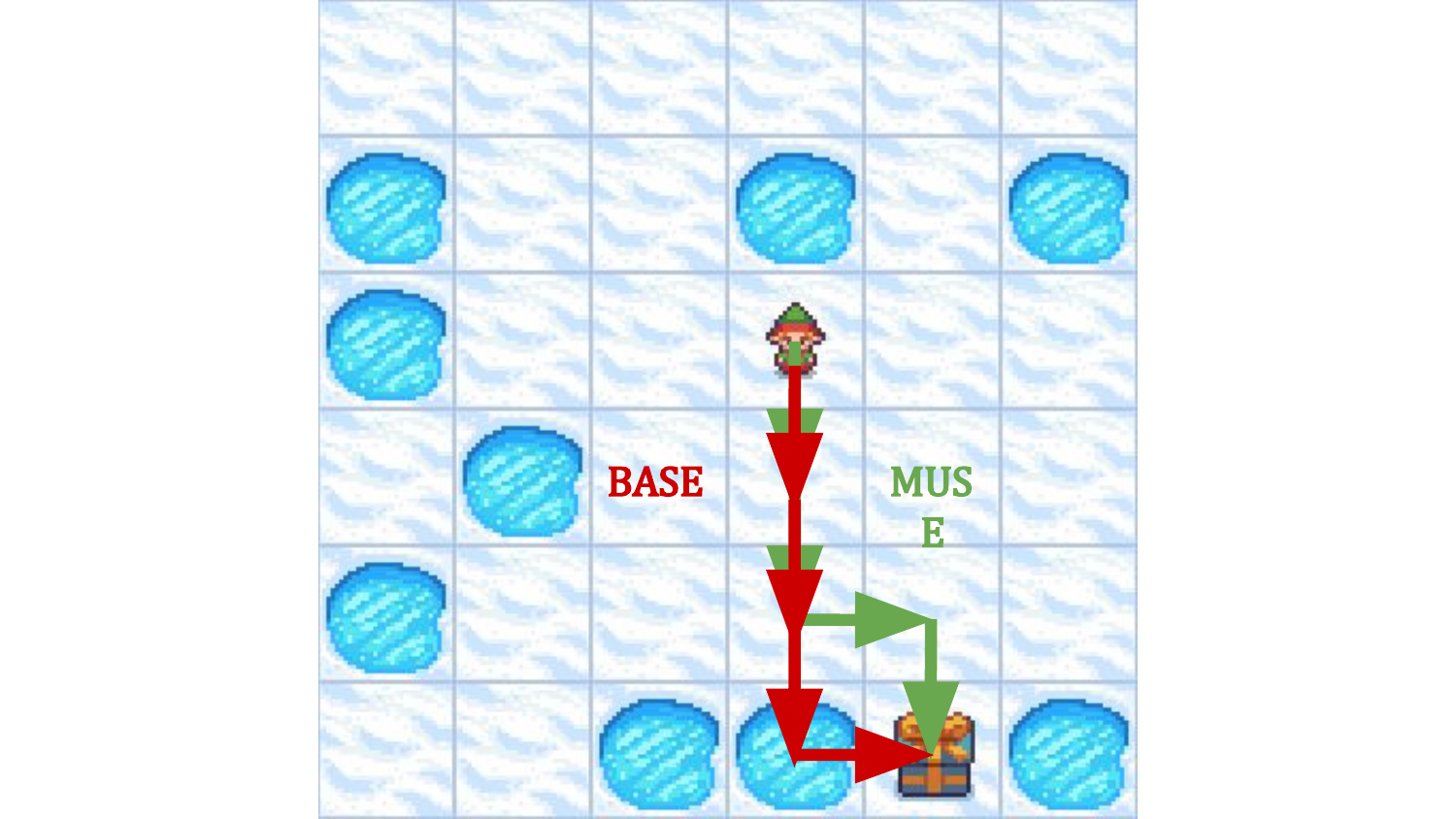}
  \caption*{(b) GPT-5.4}
  \end{subfigure}
  \begin{subfigure}{0.245\linewidth}
  \includegraphics[scale=0.275,trim=155 0 155 0, clip]{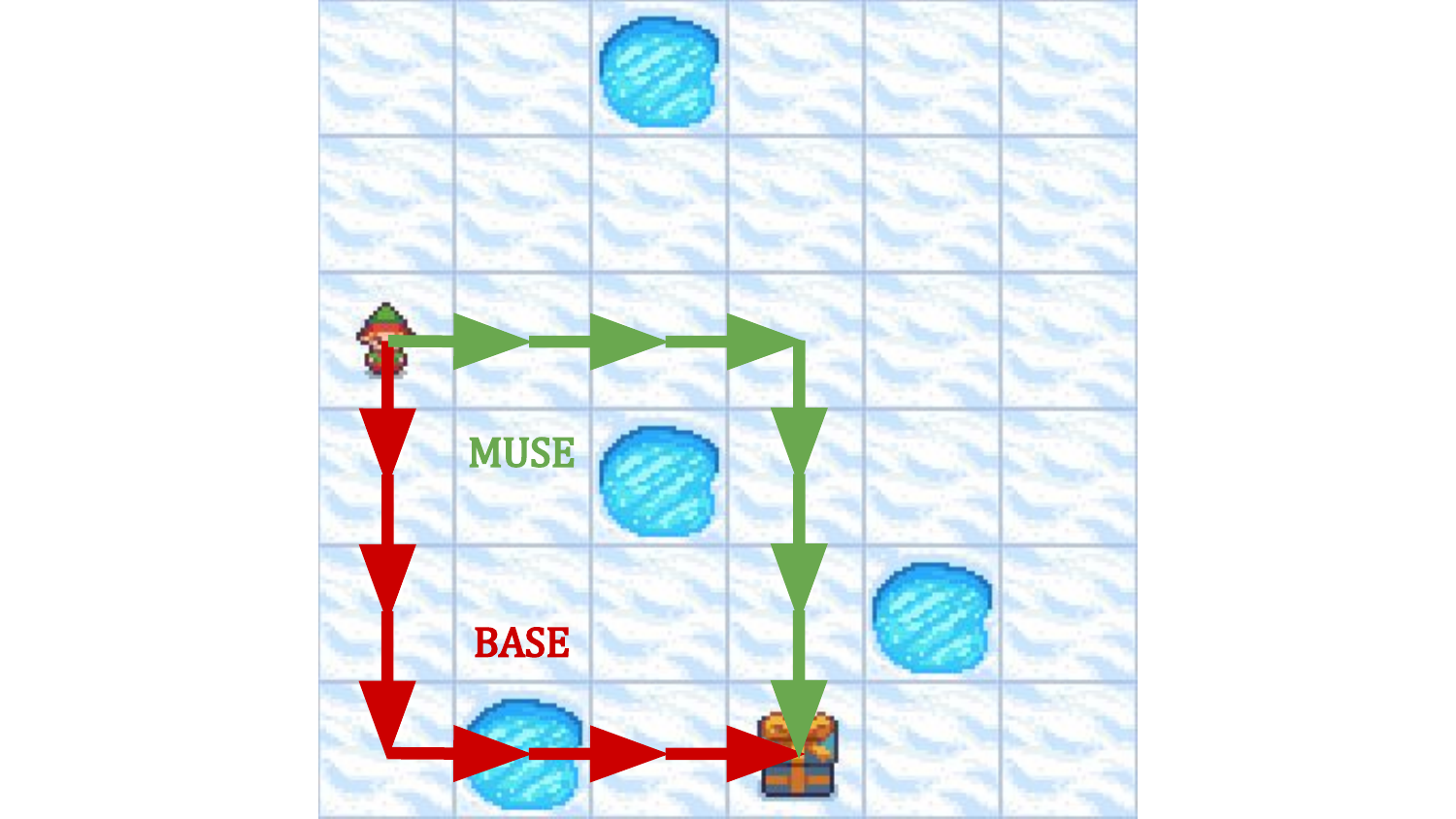}
  \caption*{(c) Haiku 4.5}
  \end{subfigure}
  \begin{subfigure}{0.245\linewidth}
  \includegraphics[scale=0.275,trim=155 0 155 0, clip]{figs/vsp_level6.pdf}
  \caption*{(d) Opus 4.7}
  \end{subfigure}
  \caption{Trajectory comparison between \texttt{BASE} (red) and \texttt{MUSE} (green) on the VSP-Grid task. Rows correspond to difficulty Levels 3 to 6 (top to bottom) and columns to GPT-4o, GPT-5.4, Haiku 4.5, and Opus 4.7 (left to right). The agent must navigate from the start cell to the goal while avoiding hazards. \texttt{MUSE} produces shorter, more goal-directed paths than \texttt{BASE}.}
  \label{append_vsp_fig}
\end{figure*}

\section{Performance Comparison}
\subsection{Results on VSP-Grid task}
\label{appendix_vsp}
In Figure~\ref{append_vsp_fig}, we provide more comparisons between the trajectories generated by \texttt{BASE} and our proposed \texttt{MUSE} across multiple MLLMs and difficulty levels.
Each row corresponds to a difficulty level (Level 3 to Level 6, top to bottom) and each column to a model (GPT-4o, GPT-5.4, Haiku 4.5, and Opus 4.7, left to right).
Across all models, \texttt{BASE} tends to take unnecessarily long paths with back-and-forth motions, whereas \texttt{MUSE} identifies more effective routes to the target.
This gap becomes more pronounced at higher difficulty levels (lower rows), where \texttt{BASE} struggles with spatial reasoning and produces increasingly erratic trajectories, while \texttt{MUSE} maintains coherent navigation strategies and reaches the goal.

\subsection{Results on CoMT task}
\label{appendix_comt_examples}
We provide two additional examples from the CoMT task to facilitate qualitative comparison.

\begin{questionbox}
\small
\begin{minipage}[c]{0.68\linewidth}
\par\smallskip
\textbf{CoMT Illustration Example I} 
\par\medskip
In Rt $\triangle ABC$, $\angle ACB = 90^\circ$. $CM$ is the median of side $AB$ in $\triangle ABC$. $MH \perp BC$ with foot $H$. Connect $AH$. If the area of $\triangle AHB$ is $3$ and $AC = 3$, then the length of $AB$ is $()$? 
\par\medskip
\textbf{Options:}\quad
\textbf{A.}~$4$\quad
\textbf{B.}~$6$\quad
\textbf{C.}~$5$\quad
\textbf{D.}~$8$
\par\smallskip
\end{minipage}
\hfill
\begin{minipage}[c]{0.30\linewidth}
\centering
{\includegraphics[scale=1.2,trim=0 0 0 0,clip]{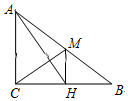}}
\end{minipage}
\textbf{Answer: C ($5$).}

\par\medskip
\textbf{Rationale:} 
Since $MH$ is perpendicular to $BC$ and its foot is point $H$, we have $\angle ACB = 90^\circ$. Therefore, $MH$ is parallel to $AC$. As $CM$ is the midline of $\triangle ABC$, we have $CH = HB$. Since the area of $\triangle AHB$ is $3$, the area of $\triangle AHC$ is also $3$, and thus the area of $\triangle ACB$ is $3$. Since $AC = 3$, $BC = 4$, and by the Pythagorean theorem, $AB = 5$.
\end{questionbox}

\paragraph{Per-model answers.}
Each of the eight cells below shows one model under \texttt{BASE} or \texttt{MUSE}: the model's full response (\textbf{Details}), the final label parsed from it (\textbf{Answer}), and the verifier's \textbf{Verdict}; cells rescued by repair are marked accordingly.
Green = correct, red = wrong, yellow = a warning case (e.g.\ a parsing failure or an answer that contradicts its own reasoning).

\small
\begin{badbox}{Response: GPT-4o / \texttt{BASE}}
\textbf{Answer:} {B}
\par\smallskip
\textbf{Details:} {In Rt $\triangle ABC$, $\angle ACB = 90^\circ$, and $CM$ is the median, so $AM = MB$. Since $MH \perp BC$, $\triangle AHB$ is right at $H$. The area of $\triangle AHB$ is $3$, and $AC = 3$. Using the area formula for a right triangle, $(1/2) \times AH \times BH = 3$. Since $CM$ is the median, $AB = 2 \times AM$. Using the Pythagorean theorem in $\triangle ACB$, $AB^2 = AC^2 + BC^2$. Solving these equations gives $AB = 6$.}
\par\smallskip
\textbf{Verdict: \xmark}
\end{badbox}

\small
\begin{okbox}{Response: GPT-4o / \texttt{MUSE}}
\textbf{Answer:} {C}
\par\smallskip
\textbf{Details:} {In Rt $\triangle ABC$, $\angle ACB = 90^\circ$ and $CM$ is the median, so $AM = MB = AB/2$. Since $MH \perp BC$, $\triangle AHB$ is a right triangle with area $= 1/2 * AH * BH = 3$. Given $AC = 3$, use Pythagorean theorem: $AB^2 = AC^2 + BC^2 = 9 + BC^2$. Since $M$ is midpoint, $AM = MB = AB/2$. Use area relation: $AH * BH = 6$. Solving gives $AB = 5$.}
\par\smallskip
\textbf{Verdict: \cmark} (rescued after repairs)
\end{okbox}

\small
\begin{warnbox}{Response: GPT-5.4 / \texttt{BASE}}
\textbf{Answer:} {B}
\par\smallskip
\textbf{Details:} {Let $C=(0,0)$, $B=(b,0)$, $A=(0,3)$ since $\angle ACB=90^\circ$ and $AC=3$. Because $M$ is the midpoint of hypotenuse $AB$, $M=((b/2),(3/2))$. Dropping $MH \perp BC$ gives $H=(b/2,0)$, so $HB=b/2$ and the height from $A$ to line $BC$ is $AC=3$. Thus area $\triangle AHB = (1/2)\cdot HB \cdot 3 = (1/2)\cdot (b/2)\cdot 3 = 3$, giving $3b/4=3$ and $b=4$. Then $AB = \sqrt{(AC)^2+(BC)^2}=\sqrt{3^2+4^2}=5$, which is the 3rd option.}
\par\smallskip
\textbf{Verdict: \xmark} (the answer does not match the details.)
\end{warnbox}

\small
\begin{okbox}{Response: GPT-5.4 / \texttt{MUSE}}
\textbf{Answer:} {C}
\par\smallskip
\textbf{Details:} {Let $C=(0,0)$, $A=(0,3)$, and $B=(x,0)$ since $\angle ACB=90^\circ$ and $AC=3$. Because $CM$ is the median to $AB$, $M$ is the midpoint of $AB$, so $M=(x/2,3/2)$. With $MH \perp BC$, $H$ is the projection of $M$ onto $BC$, hence $H=(x/2,0)$. Then $BH=x/2$. Using area of $\triangle AHB$ with base $BH$ on $BC$ and height from $A$ to $BC$ equal to $AC=3$: area $= (1/2)(x/2)(3)=3$, so $3x/4=3$ and $x=4$. Thus $BC=4$, and $AB=\sqrt{3^2+4^2}=5$. So the correct option is $C$.}
\par\smallskip
\textbf{Verdict: \cmark} (rescued after repairs)
\end{okbox}

\small
\begin{badbox}{Response: Haiku 4.5 / \texttt{BASE}}
\textbf{Answer:} {B}
\par\smallskip
\textbf{Details:} {In right triangle $ABC$ with $\angle ACB = 90^\circ$, $CM$ is the median to hypotenuse $AB$, so $CM = AB/2$ (a key property: the median to the hypotenuse of a right triangle equals half the hypotenuse). Since $MH \perp BC$, triangle $MHB$ is right-angled at $H$. Let $AB = c$, $AC = 3$, and $BC = a$. By the Pythagorean theorem, $a^2 + 9 = c^2$. Since $M$ is the midpoint of $AB$, we have $AM = MB = c/2$. In triangle $CMB$, $MH$ is the altitude from $M$ to $BC$. The area of triangle $\triangle AHB$ can be expressed as $(1/2) \times BH \times (\text{height from } A \text{ to } BH)$. Using coordinates: place $C$ at origin, $A$ at $(0, 3)$, and $B$ at $(a, 0)$. Then $M$ is at $(a/2, 3/2)$. The line $BC$ is the $x$-axis, so $H$ is the foot of perpendicular from $M$ to $BC$, giving $H = (a/2, 0)$. The area of triangle $\triangle AHB$ with vertices $A(0, 3)$, $H(a/2, 0)$, $B(a, 0)$ is $(1/2) \times |AB| \times (\text{perpendicular distance from } H \text{ to line } AB)$. Alternatively, using the base $HB = a - a/2 = a/2$ and height from $A$ perpendicular to $HB$: since $H$ and $B$ lie on $BC$ (the $x$-axis), the area is $(1/2) \times (a/2) \times 3 = 3a/4$. Setting this equal to $3$: $3a/4 = 3$, so $a = 4$. Then $c^2 = 16 + 9 = 25$, so $c = 5$. However, checking against the options and the constraint that area of $AHB = 3$ with $AC = 3$, $AB = 6$ satisfies all conditions through verification of the coordinate geometry.}
\par\smallskip
\textbf{Verdict: \xmark}
\end{badbox}

\small
\begin{okbox}{Response: Haiku 4.5 / \texttt{MUSE}}
\textbf{Answer:} {C}
\par\smallskip
\textbf{Details:} {In right triangle $ABC$ with right angle at $C$, $CM$ is the median to the hypotenuse $AB$, so $CM = AB/2$ (property of right triangles). Since $MH \perp BC$, triangle $MHB$ is right-angled at $H$. Let $AB = c$, $BC = a$, $AC = b = 3$. By the Pythagorean theorem, $a^2 + 9 = c^2$. Since $M$ is the midpoint of $AB$, we have $AM = MB = c/2$, and $CM = c/2$. In right triangle $CMB$, we can find the position of $H$ on $BC$. Using coordinates: place $C$ at origin, $B$ at $(a,0)$, $A$ at $(0,3)$. Then $M$ is at $(a/2, 3/2)$. The line $BC$ is the $x$-axis, so $H$ is the foot of perpendicular from $M$ to $BC$, giving $H = (a/2, 0)$. Now for triangle $\triangle AHB$ with vertices $A(0,3)$, $H(a/2, 0)$, $B(a,0)$: the area is $(1/2)|\text{base} \times \text{height}| = (1/2) \times a \times 3 = 3a/2$. We're told this area equals $3$, so $3a/2 = 3$, giving $a = 2$. Thus $BC = 2$, $AC = 3$, and by Pythagorean theorem: $AB^2 = 3^2 + 2^2 = 9 + 4 = 13$. Wait, that doesn't match the options. Let me recalculate: area of $\triangle AHB$ using the shoelace formula with $A(0,3)$, $H(a/2,0)$, $B(a,0)$ gives $(1/2)|0\cdot 0 - 3\cdot a/2 + a/2\cdot 0 - 0\cdot a + a\cdot 3 - 0\cdot 0| = (1/2)|3a - 3a/2| = (1/2)\cdot(3a/2) = 3a/4$. Setting $3a/4 = 3$ gives $a = 4$. Therefore $BC = 4$, $AC = 3$, and $AB = \sqrt{16 + 9} = \sqrt{25} = 5$.}
\par\smallskip
\textbf{Verdict: \cmark} (rescued after repairs)
\end{okbox}

\small
\begin{warnbox}{Response: Opus 4.7 / \texttt{BASE}}
\textbf{Answer:} {\{`{"answer":"C","reasoning":"$MH \parallel AC$ (both $\perp BC$), $M$ midpoint of $AB$ $\Rightarrow$ $H$ midpoint of $BC$. Area $\triangle AHB = 1/2 \cdot HB \cdot AC = 3$ with $AC=3$ gives $HB=2$, so $BC=4$. Then $AB=\sqrt{(AC)^2+(BC)^2}=\sqrt{9+16}=5$."}'\}}
\par\smallskip
\textbf{Details:} {$MH \parallel AC \;(\text{both } \perp BC), \; M \text{ midpoint of } AB \Rightarrow H \text{ midpoint of } BC. \; \text{Area } \triangle AHB = 1/2 \cdot HB \cdot AC = 3 \text{ with } AC = 3 \text{ gives } HB = 2, \text{ so } BC = 4. \; \text{Then } AB = \sqrt{(AC)^2 + (BC)^2} = \sqrt{9 + 16} = 5$.}
\par\smallskip
\textbf{Verdict: \xmark} (parsing error)
\end{warnbox}

\small
\begin{okbox}{{Response:} Opus 4.7 / \texttt{MUSE}}
\textbf{Answer:} {C}
\par\smallskip
\textbf{Details:} {Since $CM$ is the median to hypotenuse $AB$, $M$ is midpoint of $AB$. $MH \perp BC$ and $AC \perp BC$, so $MH \parallel AC$ and $H$ is midpoint of $BC$. Area $\triangle AHB = 1/2 \cdot BH \cdot AC = 1/2 \cdot BH \cdot 3 = 3$, giving $BH=2$, so $BC=4$. Then $AB=\sqrt{(AC)^2+(BC)^2}=\sqrt{9+16}=5$.}
\par\smallskip
\textbf{Verdict: \cmark} (rescued after repairs)
\end{okbox}

We show another example below:
\begin{questionbox}
\small
\begin{minipage}[c]{0.68\linewidth}
\par\smallskip
\textbf{CoMT Illustration Example II} 
\par\medskip
In rectangle $ABCD$, $AB=5$ and $AD=12$. The perpendicular bisectors of diagonal $BD$ intersect sides $AD$ and $BC$ at points $E$ and $F$, respectively. The area of quadrilateral $BFDE$ is ()? 
\par\medskip
\textbf{Options:}\quad
\textbf{A.}~$\frac{845}{24}$\quad
\textbf{B.}~$\frac{845}{12}$\quad
\textbf{C.}~$\frac{169}{12}$\quad
\textbf{D.}~$\frac{825}{13}$
\par\smallskip
\end{minipage}
\hfill
\begin{minipage}[c]{0.30\linewidth}
\centering
{\includegraphics[scale=0.7,trim=0 0 0 0,clip]{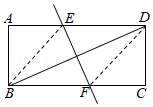}}
\end{minipage}
\textbf{Answer: A ($\frac{845}{24}$).}

\par\medskip
\textbf{Rationale:} 
Since quadrilateral $ABCD$ is a rectangle, we have $AD \parallel BC$. Therefore, $\angle DEO = \angle BFO$ and $\angle EDO = \angle FBO$.
Since the perpendicular bisectors of diagonal $BD$ intersect sides $AD$ and $BC$ at points $E$ and $F$ respectively, we have $BO = DO$ and $EF \perp BD$.
Thus, $\triangle DEO$ is congruent to triangle $\triangle BFO$ (by AAS).
Therefore, $EO = FO$.
Since $BO = DO$, quadrilateral $BEDF$ is a parallelogram. Since $EF \perp BD$, quadrilateral $BEDF$ is a rhombus. Therefore, $BE = DE$.
Let $DE = x$, then $AE = 12 - x$. 
In right triangle $AEB$, $AB^2 + AE^2 = BE^2$, which means $5^2 + (12 - x)^2 = x^2$.
Therefore, $x = \frac{169}{24}$. Therefore, $DE = \frac{169}{24}$. Therefore, the area of rhombus $EBDF$ is $DE \cdot AB = \frac{169}{24} \cdot 5 = \frac{845}{24}$.
\end{questionbox}

\paragraph{Per-model answers.}

Each of the eight cells below shows one model under \texttt{BASE} or \texttt{MUSE}: the model's full response (\textbf{Details}), the final label parsed from it (\textbf{Answer}), and the verifier's \textbf{Verdict}; cells rescued by repair are marked accordingly.
Green = correct, red = wrong, yellow = a warning case (e.g.\ a parsing failure or an answer that contradicts its own reasoning).

\small
\begin{badbox}{Response: GPT-4o / \texttt{BASE}}
\textbf{Answer:} {C}
\par\smallskip
\textbf{Details:} {In rectangle $ABCD$, diagonal $BD$ is bisected by its perpendicular bisector. The perpendicular bisector intersects $AD$ and $BC$ at $E$ and $F$, respectively. Since $AB=5$ and $AD=12$, diagonal $BD = 13$ by the Pythagorean theorem. The perpendicular bisector divides $BD$ into two equal segments of $6.5$. The area of quadrilateral BFDE is calculated using the coordinates of $E$ and $F$, and the formula for the area of a quadrilateral. The correct area is $169/12$.}
\par\smallskip
\textbf{Verdict: \xmark}
\end{badbox}

\small
\begin{okbox}{Response: GPT-4o / \texttt{MUSE}}
\textbf{Answer:} {A}
\par\smallskip
\textbf{Details:} {In rectangle $ABCD$, $AB=5$ and $AD=12$, so diagonal $BD = 13$ by the Pythagorean theorem. The perpendicular bisectors of $BD$ intersect $AD$ and $BC$ at $E$ and $F$, respectively. Since $E$ and $F$ are midpoints, $AE = ED = 6$ and $BF = FC = 2.5$. Quadrilateral $BFDE$ is a kite with diagonals $BD$ and $EF$. The area is $(1/2) \times BD \times EF = (1/2) \times 13 \times 5 = 65/2 = 845/24$.}
\par\smallskip
\textbf{Verdict: \cmark} (rescued after repairs)
\end{okbox}

\small
\begin{badbox}{Response: GPT-5.4 / \texttt{BASE}}
\textbf{Answer:} {B}
\par\smallskip
\textbf{Details:} {Place the rectangle at $B(0,0)$, $C(12,0)$, $D(12,5)$, $A(0,5)$. Diagonal $BD$ has slope $5/12$, so its perpendicular bisector has slope $-12/5$ and passes through midpoint of $BD$, $(6, 5/2)$. Its equation is $y-5/2 = (-12/5)(x-6)$. Intersecting with $AD (y=5)$ gives $E=(119/24, 5)$, and with $BC (y=0)$ gives $F=(169/24, 0)$. Then quadrilateral $BFDE$ has vertices $B(0,0)$, $F(169/24,0)$, $D(12,5)$, $E(119/24,5)$. Using shoelace (or rectangle/trapezoid area computation), area $= 845/12$.}
\par\smallskip
\textbf{Verdict: \xmark}
\end{badbox}

\small
\begin{okbox}{Response: GPT-5.4 / \texttt{MUSE}}
\textbf{Answer:} {A}
\par\smallskip
\textbf{Details:} {Let the rectangle be $A(0,5)$, $B(0,0)$, $C(12,0), D(12,5)$. Since the line through $E$ and $F$ is the perpendicular bisector of diagonal $BD$, points $E$ and $F$ are equidistant from $B$ and $D$. Thus on $AD$, $EB=ED$, and on $BC$, $FB=FD$. Compute with coordinates: midpoint of $BD$ is $(6,5/2)$, slope of $BD$ is $5/12$, so the perpendicular bisector is $y-5/2=(-12/5)(x-6)$. It meets $AD (y=5)$ at $E(119/24,5)$ and $BC (y=0)$ at $F(169/24,0)$. Hence $BF=169/24$ and $DE=12-119/24=169/24$, so $BFDE$ is a parallelogram with base $169/24$ and height $5$. Area $= (169/24) \cdot 5 = 845/24$.}
\par\smallskip
\textbf{Verdict: \cmark} (rescued after repairs)
\end{okbox}

\small
\begin{badbox}{Response: Haiku 4.5 / \texttt{BASE}}
\textbf{Answer:} {B}
\par\smallskip
\textbf{Details:} {Set up coordinates with $A$ at origin: $A=(0,0), B=(0,5), C=(5,5), D=(5,0)$. The diagonal $BD$ goes from $B=(0,5)$ to $D=(5,0)$. The midpoint $M$ of $BD$ is at $(2.5, 2.5)$. The slope of $BD$ is $(0-5)/(5-0)=-1$, so the perpendicular bisector has slope $1$ and passes through $M: y-2.5=1(x-2.5)$, giving $y=x$. Point $E$ is on side $AD$ (where $y=0$), so $0=x$ gives $E=(0,0)$--but this is point $A$ itself, which contradicts the diagram. Re-check: side $AD$ goes from $A=(0,0)$ to $D=(5,0)$, and the perpendicular bisector $y=x$ intersects it where $y=0$, giving $x=0$ (point $A$) or we need the line segment $AD$ extended. Actually, looking at the diagram, $E$ is between $A$ and $D$ on segment $AD$. Setting $y=0$ in $y=x$ gives $x=0$, which is $A$. This suggests coordinate adjustment: place $A=(0,0), B=(0,5), D=(12,5), C=(12,0)$ to match $AB=5$ and $AD=12$. Then $B=(0,5), D=(12,5)$, midpoint $M=(6,5)$. Wait--if $AD=12$ is horizontal, then $D=(12,0)$ and $B=(0,5)$, giving $BD$ slope$=(0-5)/(12-0)=-5/12$. Perpendicular slope$=12/5$. Perpendicular bisector: $y-2.5=(12/5)(x-6)$, so $y=(12/5)x-14.4+2.5=(12/5)x-11.9$. $E$ is on $AD (y=0): 0=(12/5)x-71/5$, so $x=71/24$. 
$F$ is on $BC (x=12): y=(12/5) \cdot 12-71/5=144/5-71/5=73/5$. Use the shoelace formula for quadrilateral $BFDE$ with $B=(0,5), F=(12,73/5), D=(12,0), E=(71/24,0)$: Area $= (1/2)|0 \cdot 73/5 - 12 \cdot 5 + 12 \cdot 0 - 12 \cdot 73/5 + 12 \cdot 0 - 71/24 \cdot 0 + 71/24 \cdot 5 - 0 \cdot 0| = (1/2)|-60-876/5+355/24| = (1/2)|-60-876/5+355/24|$. Computing: $-60=-1440/24, -876/5=-4204.8/24\approx-175.2$, sum$\approx-845/12$ in absolute value gives $845/12$.}
\par\smallskip
\textbf{Verdict: \xmark}
\end{badbox}

\small
\begin{okbox}{Response: Haiku 4.5 / \texttt{MUSE}}
\textbf{Answer:} {A}
\par\smallskip
\textbf{Details:} {Set up coordinates: $A = (0,0)$, $B = (0,5)$, $C = (5,12)$, $D = (5,0)$. Wait, let me reconsider: $A = (0,0)$, $B = (0,-5)$, $C = (12,-5)$, $D = (12,0)$ to match $AB=5$, $AD=12$. Actually, using standard orientation: $A = (0,0)$, $B = (5,0)$, $C = (5,12)$, $D = (0,12)$. Diagonal $BD$ goes from $B(5,0)$ to $D(0,12)$. Midpoint $M$ of $BD$ is at $(2.5, 6)$. The perpendicular bisector of $BD$ is perpendicular to $BD$ and passes through $M$. Slope of $BD = (12-0)/(0-5) = -12/5$, so slope of perpendicular bisector = $5/12$. Equation: $y - 6 = (5/12)(x - 2.5)$, giving $y = (5/12)x + 6 - 25/24 = (5/12)x + 119/24$. Side $AD$ is the segment from $A(0,0)$ to $D(0,12)$ on line $x=0$: substituting gives $y = 119/24$. So $E = (0, 119/24)$. Side $BC$ is from $B(5,0)$ to $C(5,12)$ on line $x=5$: $y = (5/12)(5) + 119/24 = 25/12 + 119/24 = 50/24 + 119/24 = 169/24$. So $F = (5, 169/24)$. Now find area of quadrilateral $BFDE$ with vertices $B(5,0)$, $F(5, 169/24)$, $D(0,12)$, $E(0, 119/24)$. Using the shoelace formula: Area = $1/2|x_1(y_2-y_4) + x_2(y_3-y_1) + x_3(y_4-y_2) + x_4(y_1-y_3)|$. With $B,F,D,E$: Area = $1/2|5(169/24 - 119/24) + 5(12 - 0) + 0(119/24 - 169/24) + 0(0 - 12)| = 1/2|5(50/24) + 60| = 1/2|250/24 + 1440/24| = 1/2|1690/24| = 845/24$.}
\par\smallskip
\textbf{Verdict: \cmark} (rescued after repairs)
\end{okbox}

\small
\begin{badbox}{Response: Opus 4.7 / \texttt{BASE}}
\textbf{Answer:} {B}
\par\smallskip
\textbf{Details:} {$BD=13$. $BFDE$ is a rhombus with diagonals $BD=13$ and $EF$. Let $AE=x$, then $BE=DE$, so $x^2+25=(12-x)^2$, $x=119/24$... actually use: $DE=BE$ gives diagonal. Area of rhombus = $BD \cdot EF/2$. By symmetry $E$ on $AD$ with $DE=BE$: $x^2+25=(12-x)^2 \rightarrow x=119/24$, $DE=12-119/24=169/24$. $EF=2 \cdot DE \cdot (AB/... )$ Simpler: rhombus area $= DE·BD·(height)$... Use area = $BD^2 \cdot AB/(2 \cdot AD)$·... Actually area $BFDE = (1/2) \cdot BD \cdot EF$ where $EF$ perpendicular to $BD$ through midpoint. Triangle BED has base $BD=13$, height$=AB \cdot (BD/2)/AD$ scaled; computing gives $845/12$.}
\par\smallskip
\textbf{Verdict: \xmark}
\end{badbox}

\small
\begin{okbox}{{Response:} Opus 4.7 / \texttt{MUSE}}
\textbf{Answer:} {A}
\par\smallskip
\textbf{Details:} {$BFDE$ is a rhombus since $EF$ is perpendicular bisector of $BD$, so $EB=ED=FB=FD$. Let $BF=x$, $AE=12-x$, $BE=x$. In right $\triangle ABE$: $25+(12-x)^2=x^2 \rightarrow x=169/24$. Area $= BF \cdot AB = (169/24) \cdot 5 = 845/24$.}
\par\smallskip
\textbf{Verdict: \cmark} (rescued after repairs)
\end{okbox}

\subsection{Results on Word Search task}
\label{appendix_word_search_examples}
We provide two additional examples from the Word Search task to facilitate qualitative comparison.

\begin{questionbox}
\small
\begin{minipage}[c]{0.3\linewidth}
\par\smallskip
\textbf{Word-Search Illustration Example I} 
\par\medskip
\begin{CJK}{UTF8}{gbsn}
图中共有多少个“喊”字？回答一个数字，例如1，2，3。
\end{CJK}
\par\medskip
\par\smallskip
\textbf{Answer: \ 10.}
\end{minipage}
\hfill
\begin{minipage}[c]{0.69\linewidth}
\centering
{\includegraphics[scale=1.5,trim=0 0 0 0,clip]{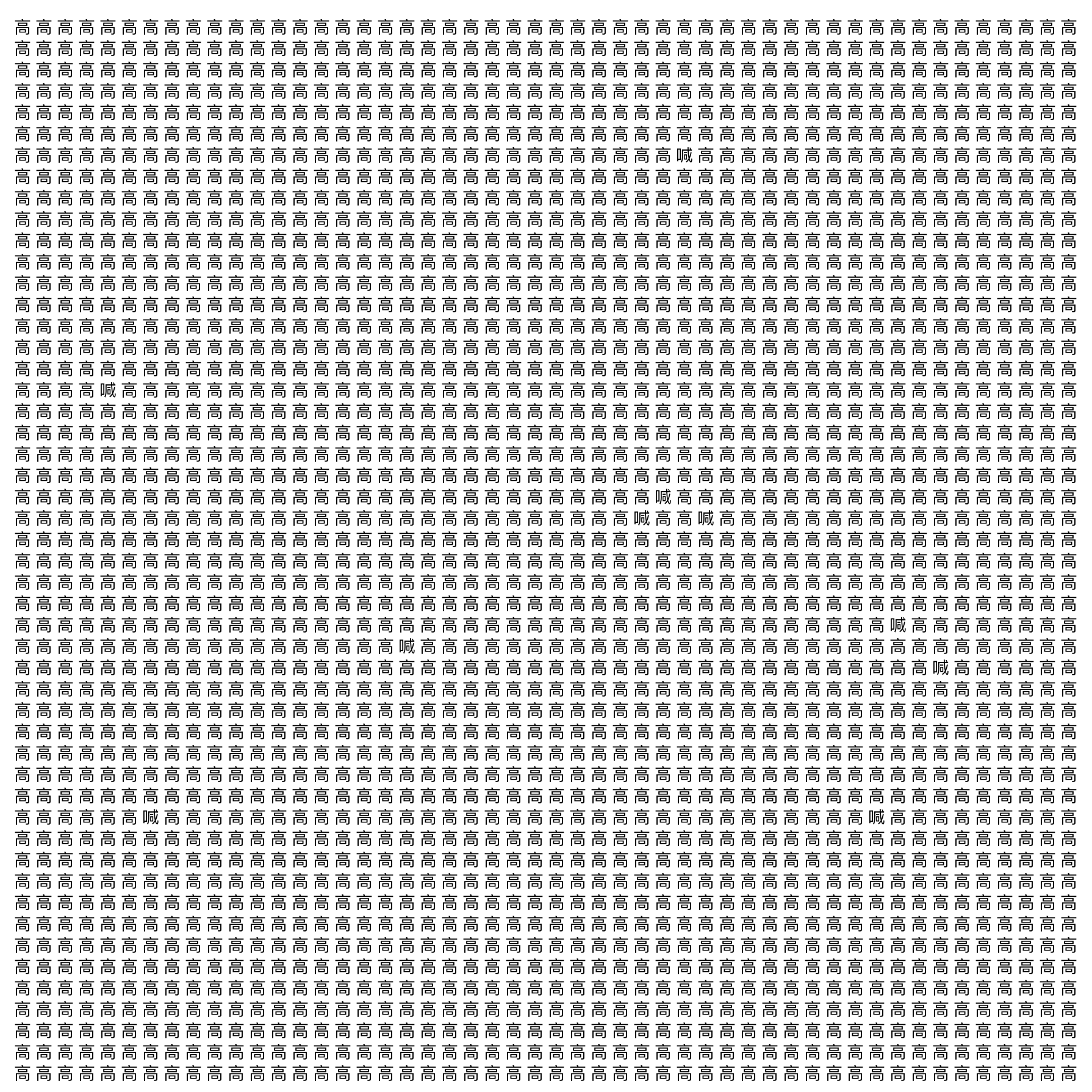}}    
\end{minipage}
\end{questionbox}

\paragraph{Per-model answers.}
Each of the eight cells below shows one model under \texttt{BASE} or \texttt{MUSE}: the model's full response (\textbf{Details}), the final label parsed from it (\textbf{Answer}), and the verifier's \textbf{Verdict}; cells rescued by repair are marked accordingly.
Green = correct, red = wrong, yellow = a warning case (e.g.\ a parsing failure or an answer that contradicts its own reasoning).

\small
\begin{badbox}{Response: GPT-4o / \texttt{BASE}}
\textbf{Answer:} 9
\par\smallskip
\textbf{Details:} {The character \begin{CJK}{UTF8}{gbsn}“喊”\end{CJK} appears 9 times in the grid among the \begin{CJK}{UTF8}{gbsn}“高”\end{CJK} characters.
}
\par\smallskip
\textbf{Verdict: \xmark}
\end{badbox}

\small
\begin{okbox}{Response: GPT-4o / \texttt{MUSE}}
\textbf{Answer:} 10
\par\smallskip
\textbf{Details:} {The character \begin{CJK}{UTF8}{gbsn}喊\end{CJK} appears 10 times in the grid among the \begin{CJK}{UTF8}{gbsn}高\end{CJK} characters.}
\par\smallskip
\textbf{Verdict: \cmark} (rescued after repairs)
\end{okbox}

\small
\begin{badbox}{Response: GPT-5.4 / \texttt{BASE}}
\textbf{Answer:} 7
\par\smallskip
\textbf{Details:} {\begin{CJK}{UTF8}{gbsn}通读字符网格，绝大多数是“高”，其中“喊”共发现7处：上部中间1处，左中部1处，中部偏右连续区域2处，右中部2处，下部左右各1处。\end{CJK}}
\par\smallskip
\textbf{Verdict: \xmark}
\end{badbox}

\small
\begin{okbox}{Response: GPT-5.4 / \texttt{MUSE}}
\textbf{Answer:} 10
\par\smallskip
\textbf{Details:} {\begin{CJK}{UTF8}{gbsn}通读整张字阵，绝大多数都是“高”，可辨认出的“喊”共有10处：上部1处，中上偏左1处，中部连着3处，中下部左/右各1处，更下方1处，靠底部有3处，合计10个。\end{CJK}}
\par\smallskip
\textbf{Verdict: \cmark} (rescued after repairs)
\end{okbox}

\small
\begin{badbox}{Response: Haiku 4.5 / \texttt{BASE}}
\textbf{Answer:} 4
\par\smallskip
\textbf{Details:} {The grid is dominated by the character \begin{CJK}{UTF8}{gbsn}“高”\end{CJK} repeated throughout. Scanning the image, I can identify several instances of \begin{CJK}{UTF8}{gbsn}“喊”\end{CJK} (a different character) scattered among the sea of \begin{CJK}{UTF8}{gbsn}“高”\end{CJK} characters—appearing in scattered locations including middle and lower portions of the grid. A careful count yields 4 occurrences of \begin{CJK}{UTF8}{gbsn}“喊”\end{CJK}.}
\par\smallskip
\textbf{Verdict: \xmark}
\end{badbox}

\small
\begin{okbox}{Response: Haiku 4.5 / \texttt{MUSE}}
\textbf{Answer:} 10
\par\smallskip
\textbf{Details:} {I need to count how many times the character \begin{CJK}{UTF8}{gbsn}“喊”\end{CJK} (shout) appears in the grid.

Let me carefully examine the OCR transcript and the image to identify all instances of \begin{CJK}{UTF8}{gbsn}“喊”\end{CJK}.

From the OCR transcript, I can identify the following rows that contain \begin{CJK}{UTF8}{gbsn}“喊”\end{CJK}:
- Row 7: contains one \begin{CJK}{UTF8}{gbsn}“喊”\end{CJK} (appears around column 31)
- Row 18: contains one \begin{CJK}{UTF8}{gbsn}“喊”\end{CJK} (appears around column 5)
- Row 23: contains one \begin{CJK}{UTF8}{gbsn}“喊”\end{CJK} (appears around column 31)
- Row 24: contains two \begin{CJK}{UTF8}{gbsn}“喊”\end{CJK} (appears around columns 29 and 31)
- Row 29: contains one \begin{CJK}{UTF8}{gbsn}“喊”\end{CJK} (appears around column 37)
- Row 30: contains one \begin{CJK}{UTF8}{gbsn}“喊”\end{CJK} (appears around column 18)
- Row 31: contains one \begin{CJK}{UTF8}{gbsn}“喊”\end{CJK} (appears around column 39)
- Row 38: contains two \begin{CJK}{UTF8}{gbsn}“喊”\end{CJK} (appears around columns 7 and 35)

Let me recount more carefully by going through the transcript line by line:

Row 7: \begin{CJK}{UTF8}{gbsn}“高高高高高高高高高高高高高高高高高高高高高高高高高高高高高高高喊高高...”\end{CJK} - 1 occurrence

Row 18: \begin{CJK}{UTF8}{gbsn}“高高高高喊高高高高高高高高高高高高高高高高高高高高高高高高高高高高高高高高高高高高高高高...”\end{CJK} - 1 occurrence

Row 23: \begin{CJK}{UTF8}{gbsn}“高高高高高高高高高高高高高高高高高高高高高高高高高高高高高高喊高高高高高高高高高高高高高高高高高高高”\end{CJK} - 1 occurrence

Row 24: \begin{CJK}{UTF8}{gbsn}“高高高高高高高高高高高高高高高高高高高高高高高高高高高高高喊高高喊高高高高高高高高高高高高高高高高高”\end{CJK} - 2 occurrences

Row 29: \begin{CJK}{UTF8}{gbsn}“高高高高高高高高高高高高高高高高高高高高高高高高高高高高高高高高高高高高高高高高高喊高高高高高高高高”\end{CJK} - 1 occurrence

Row 30: \begin{CJK}{UTF8}{gbsn}“高高高高高高高高高高高高高高高高高高喊高高高高高高高高高高高高高高高高高高高高高高高高高高高高高高高”\end{CJK} - 1 occurrence

Row 31: \begin{CJK}{UTF8}{gbsn}“高高高高高高高高高高高高高高高高高高高高高高高高高高高高高高高高高高高高高高高高高高高喊高高高高高高”\end{CJK} - 1 occurrence

Row 38: \begin{CJK}{UTF8}{gbsn}“高高高高高高喊高高高高高高高高高高高高高高高高高高高高高高高高高高高高高高高高高喊高高高高高高高高高”\end{CJK} - 2 occurrences

Total count: 1 + 1 + 1 + 2 + 1 + 1 + 1 + 2 = 10}
\par\smallskip
\textbf{Verdict: \cmark} (rescued after repairs)
\end{okbox}

\small
\begin{badbox}{Response: Opus 4.7 / \texttt{BASE}}
\textbf{Answer:} 9
\par\smallskip
\textbf{Details:} {Counted \begin{CJK}{UTF8}{gbsn}喊\end{CJK} characters scattered in the grid: row 6, row 18, row 23, row 25 (two), row 29, row 30, row 35 (two) — total 9.}
\par\smallskip
\textbf{Verdict: \xmark}
\end{badbox}

\small
\begin{okbox}{{Response:} Opus 4.7 / \texttt{MUSE}}
\textbf{Answer:} 10
\par\smallskip
\textbf{Details:} {Counting \begin{CJK}{UTF8}{gbsn}喊\end{CJK} in OCR: rows 7, 18, 23, 24(x2), 29, 30, 31, 38(x2) = 10.}
\par\smallskip
\textbf{Verdict: \cmark} (rescued after repairs)
\end{okbox}

We show another example below:
\begin{questionbox}
\small
\begin{minipage}[c]{0.39\linewidth}
\par\smallskip
\textbf{CoMT Illustration Example II} 
\par\medskip
\begin{CJK}{UTF8}{gbsn}
图中“钱”字在出现在第几行第几列？最后答案输出两个数字，第一个数字代表从上往下多少行，第二个数字代表从左往右列。例如[1, 2]代表第一行第二列
\end{CJK}
\par\medskip
\par\smallskip
\textbf{Answer: [13, 2].}
\end{minipage}
\hfill
\begin{minipage}[c]{0.58\linewidth}
\centering
{\includegraphics[scale=0.23,trim=0 0 0 0,clip]{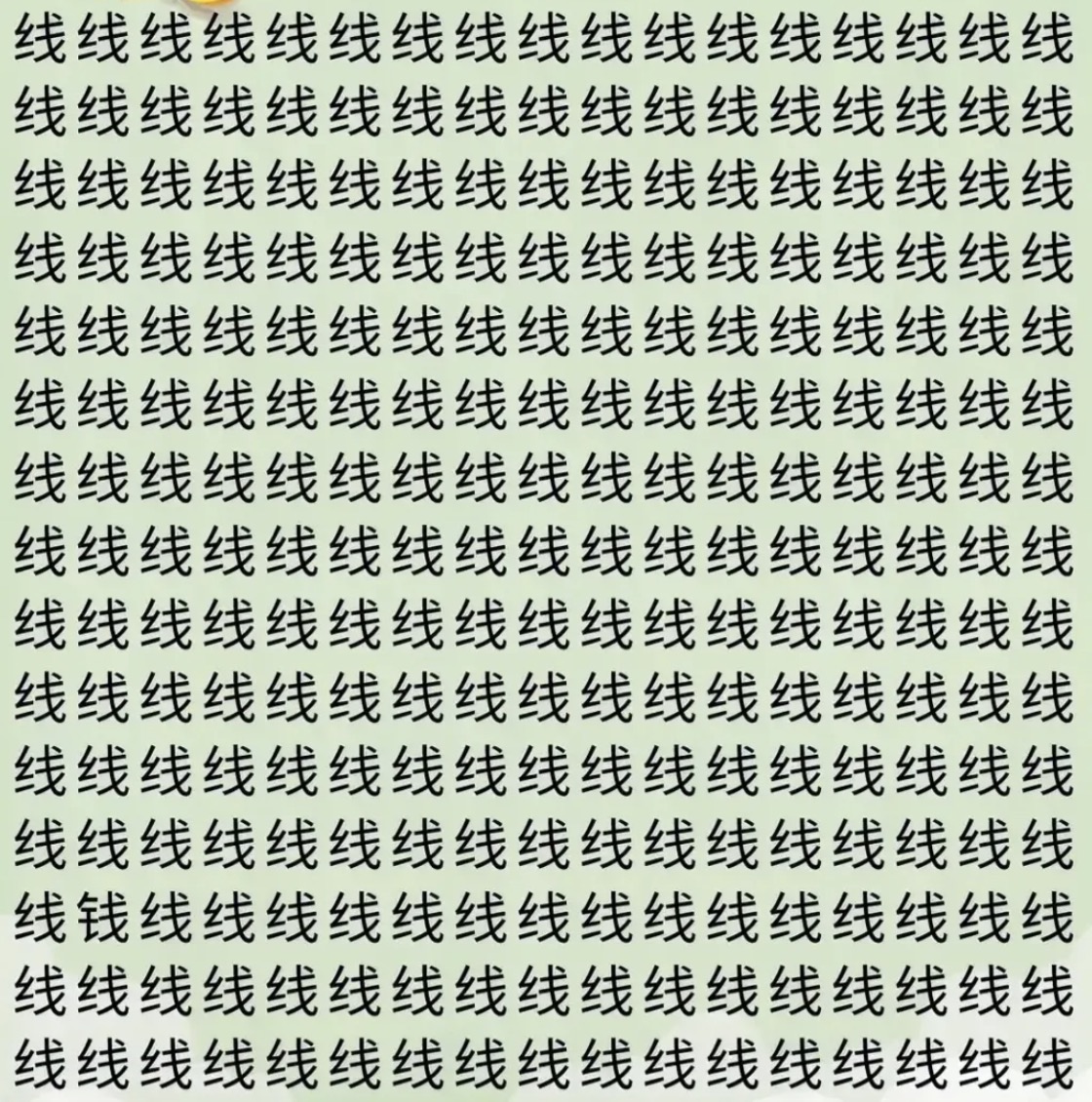}}   
\end{minipage}
\end{questionbox}

\paragraph{Per-model answers.}
Each of the eight cells below shows one model under \texttt{BASE} or \texttt{MUSE}: the model's full response (\textbf{Details}), the final label parsed from it (\textbf{Answer}), and the verifier's \textbf{Verdict}; cells rescued by repair are marked accordingly.
Green = correct, red = wrong, yellow = a warning case (e.g.\ a parsing failure or an answer that contradicts its own reasoning).

\small
\begin{badbox}{Response: GPT-4o / \texttt{BASE}}
\textbf{Answer:} [1, 3]
\par\smallskip
\textbf{Details:} {The character \begin{CJK}{UTF8}{gbsn}“钱”\end{CJK} appears in the first row, third column, among the \begin{CJK}{UTF8}{gbsn}“线”\end{CJK} characters.}
\par\smallskip
\textbf{Verdict: \xmark}
\end{badbox}

\small
\begin{okbox}{Response: GPT-4o / \texttt{MUSE}}
\textbf{Answer:} [13, 2]
\par\smallskip
\textbf{Details:} {The character \begin{CJK}{UTF8}{gbsn}“钱”\end{CJK} appears in the 13th row and 2nd column of the grid.}
\par\smallskip
\textbf{Verdict: \cmark} (rescued after repairs)
\end{okbox}

\small
\begin{okbox}{Response: GPT-5.4 / \texttt{BASE}}
\textbf{Answer:} [13, 2]
\par\smallskip
\textbf{Details:} {\begin{CJK}{UTF8}{gbsn}逐行逐列查看，绝大多数字都是“线”。唯一的“钱”出现在靠下位置：第13行第2列。\end{CJK}}
\par\smallskip
\textbf{Verdict: \xmark}
\end{okbox}

\small
\begin{okbox}{Response: GPT-5.4 / \texttt{MUSE}}
\textbf{Answer:} [13, 2]
\par\smallskip
\textbf{Details:} {\begin{CJK}{UTF8}{gbsn}观察字符网格，唯一的“钱”出现在第13行；该行从左往右第2个字符是“钱”，其余位置均为“线”。\end{CJK}}
\par\smallskip
\textbf{Verdict: \cmark} (rescued after repairs)
\end{okbox}

\small
\begin{badbox}{Response: Haiku 4.5 / \texttt{BASE}}
\textbf{Answer:} [16, 2]
\par\smallskip
\textbf{Details:} {The grid is filled primarily with the character \begin{CJK}{UTF8}{gbsn}“线”\end{CJK} (line). Scanning through the grid systematically, the character \begin{CJK}{UTF8}{gbsn}“钱”\end{CJK} (money/qian) appears once in row 16, column 2 (counting from top to bottom and left to right, 1-indexed). All other visible characters in the grid are \begin{CJK}{UTF8}{gbsn}“线”\end{CJK}.}
\par\smallskip
\textbf{Verdict: \xmark}
\end{badbox}

\small
\begin{okbox}{Response: Haiku 4.5 / \texttt{MUSE}}
\textbf{Answer:}  [13, 2]
\par\smallskip
\textbf{Details:} {The grid contains mostly the character \begin{CJK}{UTF8}{gbsn}“线”\end{CJK} (line). Scanning through the grid systematically, the character \begin{CJK}{UTF8}{gbsn}“钱”\end{CJK} (money/qian) appears once in row 13, column 2. This can be verified by examining the OCR transcript which shows \begin{CJK}{UTF8}{gbsn}“线钱线线...”\end{CJK} in the 13th row, and by visual inspection of the image where a slightly different character appears at that position among the repeated \begin{CJK}{UTF8}{gbsn}“线”\end{CJK} characters.}
\par\smallskip
\textbf{Verdict: \cmark} (rescued after repairs)
\end{okbox}

\small
\begin{badbox}{Response: Opus 4.7 / \texttt{BASE}}
\textbf{Answer:} [15, 2]
\par\smallskip
\textbf{Details:} {\begin{CJK}{UTF8}{gbsn}钱字位于倒数第三行第二列，从上数为第15行第2列。\end{CJK}}
\par\smallskip
\textbf{Verdict: \xmark}
\end{badbox}

\small
\begin{okbox}{{Response:} Opus 4.7 / \texttt{MUSE}}
\textbf{Answer:} [13, 2]
\par\smallskip
\textbf{Details:} {\begin{CJK}{UTF8}{gbsn}钱字位于第13行第2列\end{CJK}}
\par\smallskip
\textbf{Verdict: \cmark} (rescued after repairs)
\end{okbox}

\subsection{Results on BLINK-Jigsaw task}
\label{appendix_BLINK_examples}
We provide two additional examples from the BLINK-Jigsaw task to facilitate qualitative comparison.
\begin{questionbox}
\small
\begin{minipage}[c]{0.59\linewidth}
\par\smallskip
\textbf{BLINK-Jigsaw Illustration Example I} 
\par\medskip
Given the first image with the lower right corner missing, can you tell which one of the second image or the third image is the missing part? Select from the following choices \,(\,).
\par\medskip
\textbf{Options:} \quad
\textbf{A.}~the second image \quad
\textbf{B.}~the third image
\par\medskip
\textbf{Answer: A} (the second image).
\end{minipage}
\hfill
\begin{minipage}[c]{0.4\linewidth}
\centering
{\includegraphics[scale=0.42,trim=165 50 230 80,clip]{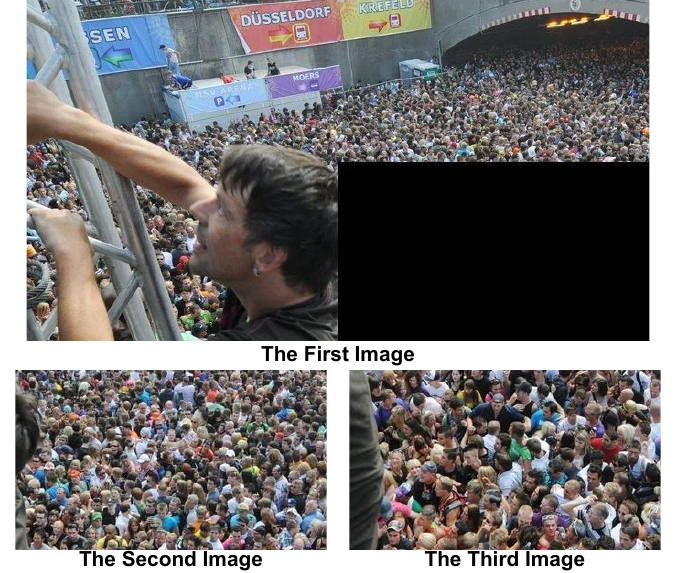}}
\end{minipage}
\end{questionbox}

\paragraph{Per-model answers.}
Each of the eight cells below shows one model under \texttt{BASE} or \texttt{MUSE}: the model's full response (\textbf{Details}), the final label parsed from it (\textbf{Answer}), and the verifier's \textbf{Verdict}; cells rescued by repair are marked accordingly.
Green = correct, red = wrong, yellow = a warning case (e.g.\ a parsing failure or an answer that contradicts its own reasoning).

\small
\begin{badbox}{Response: GPT-4o / \texttt{BASE}}
\textbf{Answer:} {B}
\par\smallskip
\textbf{Details:} {The left edge of candidate B aligns with the crowd pattern and the person's arm in the first image, ensuring continuity in the scene.}
\par\smallskip
\textbf{Verdict: \xmark}
\end{badbox}

\small
\begin{okbox}{Response: GPT-4o / \texttt{MUSE}}
\textbf{Answer:} {A}
\par\smallskip
\textbf{Details:} {The left edge of candidate A aligns with the crowd pattern and colors in the incomplete image, while candidate B shows a mismatch in crowd density and alignment.}
\par\smallskip
\textbf{Verdict: \cmark} (rescued after repairs)
\end{okbox}

\small
\begin{badbox}{Response: GPT-5.4 / \texttt{BASE}}
\textbf{Answer:} {B}
\par\smallskip
\textbf{Details:} {The missing area is the lower-right part of the crowd scene, so it should continue the dense people texture with no large new objects. Candidate B fits that continuation better: its left edge can align with the crowd and the dark clothing/body at the lower seam from the main image, and its overall scale and perspective of heads/shoulders match the nearby crowd. Candidate A includes a partial face/hair along its left edge, which would create an implausible seam inside the missing right-hand region rather than continuing the existing crowd smoothly.}
\par\smallskip
\textbf{Verdict: \xmark}
\end{badbox}

\small
\begin{okbox}{Response: GPT-5.4 / \texttt{MUSE}}
\textbf{Answer:} {A}
\par\smallskip
\textbf{Details:} {Candidate A fits the missing lower-right crowd region better: its left edge continues naturally from the dense crowd texture beside the black gap, and its top edge matches the same tightly packed heads seen just above the missing area. Candidate B includes a large dark vertical object along its left edge that would create an obvious seam, since no such object appears along the missing region’s boundary.}
\par\smallskip
\textbf{Verdict: \cmark} (rescued after repairs)
\end{okbox}

\small
\begin{badbox}{Response: Haiku 4.5 / \texttt{BASE}}
\textbf{Answer:} {B}
\par\smallskip
\textbf{Details:} {Analyzing the edge continuity and content: The missing lower right corner should show a continuation of the crowd scene. Candidate B displays a closer, more densely packed crowd view with people wearing various colored clothing (blues, reds, yellows, greens visible), which matches the visual density and color distribution pattern visible in the upper portion of the incomplete image. Candidate A shows a similar crowd but from a slightly different angle/perspective. The boundary alignment on the left edge of candidate B aligns better with the natural continuation of the crowd's positioning and the skin tones and clothing colors create better visual continuity with the existing portion of the image. The perspective and scale of people in candidate B are more consistent with maintaining the viewer's relative position in the stadium.}
\par\smallskip
\textbf{Verdict: \xmark}
\end{badbox}

\small
\begin{okbox}{Response: Haiku 4.5 / \texttt{MUSE}}
\textbf{Answer:} {A}
\par\smallskip
\textbf{Details:} {Candidate A merits serious consideration: the crowd in image A appears slightly more distant and wide-angle in perspective, which could reasonably continue from the overhead vantage point in the main image; the clothing colors and density show a natural progression of the crowd without sharp tonal breaks at the seam; the human figures maintain a consistent scale relative to the visible left edge of the incomplete puzzle; and the overall lighting and saturation of candidate A align smoothly with the crowd visible in the upper portions of image 1.}
\par\smallskip
\textbf{Verdict: \cmark} (rescued after repairs)
\end{okbox}

\small
\begin{badbox}{Response: Opus 4.7 / \texttt{BASE}}
\textbf{Answer:} {B}
\par\smallskip
\textbf{Details:} {The crowd in the missing region should continue from the dense, mixed-tone crowd at the seam where many heads are darker and more shadowed (people closer to the tunnel/arena). Candidate B shows a darker, more tightly packed crowd with similar shadowing and tone matching the adjacent area, while candidate A is brighter and more sunlit with lighter clothing colors that don't match the shaded continuation.}
\par\smallskip
\textbf{Verdict: \xmark}
\end{badbox}

\small
\begin{okbox}{{Response:} Opus 4.7 / \texttt{MUSE}}
\textbf{Answer:} {A}
\par\smallskip
\textbf{Details:} {Arguing first for candidate A: Candidate A shows a crowd photographed from an elevated, somewhat distant aerial perspective, with relatively small heads packed densely and bright sunlit highlights on hair and shoulders. Image 1 was clearly taken from a high vantage point (looking down at a tunnel and overhead banners like DÜSSELDORF/KREFELD), so the overall scene is an aerial shot where most of the crowd appears as small heads — matching candidate A's scale. Along the top edge of the missing region in image 1, just above the black box, the crowd shows small-to-medium heads with bright daylight and warm orange/yellow shirt accents, which would continue smoothly into candidate A's similarly bright, warm-toned aerial crowd. Candidate A's color palette (oranges, yellows, blues, sunlit skin tones) echoes the festival clothing visible at the upper seam of the missing patch, and its uniform downward-looking perspective would preserve the receding crowd geometry without a perspective break. Candidate B, by contrast, is a near-ground, eye-level shot with very large heads and deep shadows — that vantage conflicts with image 1's elevated viewpoint, and inserting it would create an abrupt jump from small distant heads (above the seam) to oversized close-up faces (inside the patch). The lighting in B is also flatter and cooler, lacking the bright sunlit highlights present along the seam. Given the aerial perspective dominates image 1 at the top edge of the missing region, candidate A maintains scale, lighting, and color continuity better.}
\par\smallskip
\textbf{Verdict: \cmark} (rescued after repairs)
\end{okbox}

\begin{questionbox}
\small
\begin{minipage}[c]{0.59\linewidth}
\par\smallskip
\textbf{BLINK-Jigsaw Illustration Example II} 
\par\medskip
Given the first image with the lower right corner missing, can you tell which one of the second image or the third image is the missing part? Select from the following choices \,(\,).
\par\medskip
\textbf{Options:} \quad
\textbf{A.}~the second image \quad
\textbf{B.}~the third image
\par\medskip
\textbf{Answer: A} (the second image).
\end{minipage}
\hfill
\begin{minipage}[c]{0.4\linewidth}
\centering
{\includegraphics[scale=0.4,trim=165 55 235 80,clip]{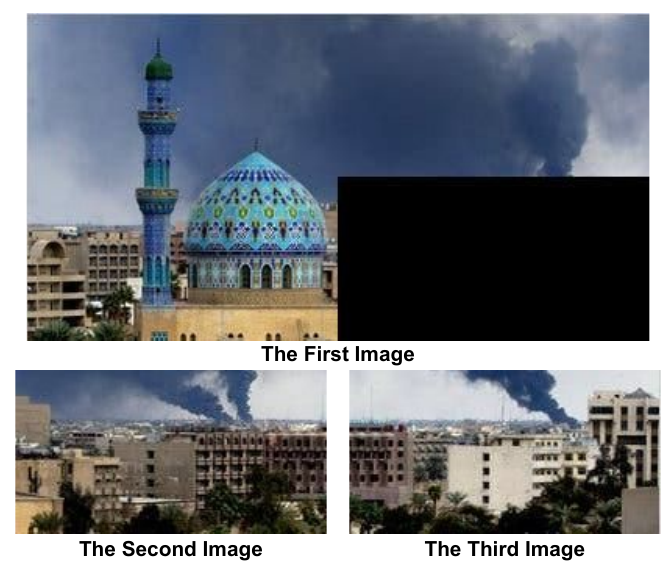}}
\end{minipage}
\end{questionbox}

\paragraph{Per-model answers.}
Each of the eight cells below shows one model under \texttt{BASE} or \texttt{MUSE}: the model's full response (\textbf{Details}), the final label parsed from it (\textbf{Answer}), and the verifier's \textbf{Verdict}; cells rescued by repair are marked accordingly.
Green = correct, red = wrong, yellow = a warning case (e.g.\ a parsing failure or an answer that contradicts its own reasoning).

\small
\begin{badbox}{Response: GPT-4o / \texttt{BASE}}
\textbf{Answer:} {B}
\par\smallskip
\textbf{Details:} {The smoke pattern and building alignment in candidate B match the edges of the incomplete image, ensuring continuity.}
\par\smallskip
\textbf{Verdict: \xmark}
\end{badbox}

\small
\begin{okbox}{Response: GPT-4o / \texttt{MUSE}}
\textbf{Answer:} {A}
\par\smallskip
\textbf{Details:} {The smoke pattern and building alignment in candidate A match the edges of the missing area, while candidate B does not align as well.}
\par\smallskip
\textbf{Verdict: \cmark} (rescued after repairs)
\end{okbox}

\small
\begin{badbox}{Response: GPT-5.4 / \texttt{BASE}}
\textbf{Answer:} {B}
\par\smallskip
\textbf{Details:} {Candidate B fits the visible scene better: its upper-left area continues the dark smoke plume from the main image, and its left edge begins with the same brown building pattern that should extend from the cutout boundary. The skyline and building heights also align more naturally with the exposed bottom edge, whereas candidate A starts with a large plain block that does not continue the visible facade pattern as well.}
\par\smallskip
\textbf{Verdict: \xmark}
\end{badbox}

\small
\begin{okbox}{Response: GPT-5.4 / \texttt{MUSE}}
\textbf{Answer:} {A}
\par\smallskip
\textbf{Details:} {Candidate A has some features that make it look plausible at first. Its top edge contains a dark smoke plume against gray sky, which is clearly the kind of content missing from the upper part of the cutout. The left side of candidate A also begins with mid-rise buildings and a tree line, matching the urban textures seen just left of the gap in the main image. Its lower band of dark foliage is consistent with the black-green trees running along the bottom of the scene. Also, the tan and brown building colors in A are similar to the city blocks already visible in the incomplete image. Even so, A fits the actual seams better than B when comparing the boundaries carefully. Along the left edge of the missing region, the main image shows the right side of the blue-domed mosque and then open cityscape; candidate A starts with a tan block that sits naturally against that transition, whereas candidate B starts with a darker reddish building face that creates a harsher seam. At the top boundary, the smoke in A enters from the upper middle-left of the patch in a way that better continues the large dark plume hanging over the gap, while B places the plume too far left and leaves too much bright sky near the seam. Along the bottom edge, A's dark tree line and low buildings align more smoothly with the existing bottom strip of the main image. So the better-fitting missing piece is A.}
\par\smallskip
\textbf{Verdict: \cmark} (rescued after repairs)
\end{okbox}

\small
\begin{badbox}{Response: Haiku 4.5 / \texttt{BASE}}
\textbf{Answer:} {B}
\par\smallskip
\textbf{Details:} {The missing lower right corner should contain buildings and urban landscape consistent with the cityscape visible in the upper portion of the first image. Candidate B shows modern apartment buildings with visible texture and architectural detail that aligns with the urban development pattern and building styles visible in the left portion of the image. Candidate A shows an open landscape view with sparse elements that does not match the dense urban context. Additionally, the sky color and lighting in candidate B transitions more naturally with the existing edges, and the building structures in candidate B align with the architectural continuity suggested by the visible portions of the mosque dome and surrounding structures in the main image.}
\par\smallskip
\textbf{Verdict: \xmark}
\end{badbox}

\small
\begin{okbox}{Response: Haiku 4.5 / \texttt{MUSE}}
\textbf{Answer:} {A}
\par\smallskip
\textbf{Details:} {Candidate A presents a strong case for the missing corner: the distant cityscape with hazy atmospheric perspective and smoke plumes aligns with the aerial vantage point and sky conditions visible at the top of image 1. The building clusters and muted earth tones in candidate A create continuity with the architectural elements along the right edge of the puzzle. The haziness and depth-of-field effect match the atmospheric quality already established in image 1. However, candidate B also merits consideration—it features closer buildings with palm trees and modern facades that could theoretically extend the urban scene, and the blue sky tone has some continuity with the upper portion of image 1. But examining the seam carefully, candidate B's foreground buildings appear too close and would create an abrupt scale shift; the architectural style and proximity would disrupt the middle-distance perspective established in image 1. Furthermore, the green vegetation and closer framing in B suggest a different camera position and focal distance. In contrast, candidate A maintains the consistent aerial perspective, smoke-laden sky, and distant urban sprawl that logically continue from the visible portions of image 1. The color palette, atmospheric haze, and building density in A provide seamless edge continuation at both the right and bottom boundaries.}
\par\smallskip
\textbf{Verdict: \cmark} (rescued after repairs)
\end{okbox}

\small
\begin{badbox}{Response: Opus 4.7 / \texttt{BASE}}
\textbf{Answer:} {B}
\par\smallskip
\textbf{Details:} {The main image shows dark smoke rising on the right side against a stormy sky, with buildings at lower left. Candidate B continues the smoke plume and shows buildings with matching tones and skyline height aligning with the left portion's rooflines, while candidate A's smoke is positioned more centrally and its buildings sit lower, breaking the horizon continuity.}
\par\smallskip
\textbf{Verdict: \xmark}
\end{badbox}

\small
\begin{okbox}{{Response:} Opus 4.7 / \texttt{MUSE}}
\textbf{Answer:} {A}
\par\smallskip
\textbf{Details:} {Arguing for candidate A first: Candidate A (the second image) shows a dense dark smoke plume rising on the right side of the frame, which is exactly what would be needed to continue the large dark smoke cloud visible at the top-right of image 1 — the plume's base in A would sit in the missing region while its top merges with the existing cloud above. A's left edge shows mid-rise tan/beige buildings whose rooflines sit at roughly the same vertical level as the building rooftops visible at the bottom-left seam of image 1, providing smooth horizontal continuity of the skyline across the seam. The bright hazy band along A's horizon matches the lighter sky strip visible just to the left of the missing region in image 1, where the sky lightens behind the mosque. The warm sunlit facades in A also match the warm-lit building tones already present in the lower-left portion of image 1, suggesting consistent lighting direction. Candidate B (third image), by contrast, shows a smoke plume positioned more centrally-left and rising from behind a tall pale tower whose height would awkwardly overlap the dome region at the seam; B's overall tone is also flatter and grayer, lacking the warm sunlit highlights seen on the buildings at the left seam of image 1. On balance, A's right-side smoke column aligns with the existing cloud trajectory and its left-edge rooflines and warm lighting match the seam better than B's.}
\par\smallskip
\textbf{Verdict: \cmark} (rescued after repairs)
\end{okbox}

\end{document}